%% file: main.tex
\documentclass[runningheads]{llncs}

\usepackage{eccv}

\usepackage{eccvabbrv}

\usepackage{graphicx}
\usepackage{booktabs}
\usepackage{amsmath,amssymb}
\usepackage{multirow}
\usepackage{makecell}
\usepackage{array}
\usepackage{enumerate}
\usepackage{pifont}
\usepackage{xcolor}
\usepackage{colortbl}
\usepackage{soul}
\usepackage{microtype}

\newcommand{\cmark}{\textcolor{red}{\ding{51}}}
\newcommand{\xmark}{\textcolor{black}{\ding{55}}}

\usepackage{wrapfig}
\usepackage{needspace}
\usepackage{siunitx}
\usepackage[normalem]{ulem}

\definecolor{mygray}{HTML}{EAF3FA}

\usepackage[accsupp]{axessibility}  

\usepackage{hyperref}

\usepackage{orcidlink}

\begin{document}

\title{\large UniMotion: A Unified Framework for Motion-Text-Vision Understanding and Generation} 

\titlerunning{UniMotion}

\author{\small { Ziyi Wang$^{1,*}$ \and Xinshun Wang$^{1,*}$ \and Shuang Chen$^{2,*}$ \and Yang Cong$^{3}$ \and Mengyuan Liu$^{1,\dagger}$}}

\authorrunning{Z.~Wang et al.}

\institute{$^{1}$State Key Laboratory of General Artificial Intelligence, \\Peking University, Shenzhen Graduate School, China \\
$^{2}$Donghua University, China\\
$^{3}$School of Automation Science and Engineering, South China University of Technology\\
\href{https://wangzy01.github.io/UniMotion}{\textcolor{blue}{https://wangzy01.github.io/UniMotion}}}

\maketitle

\begin{center}
\vspace{-6pt}
    \begin{minipage}{0.8\linewidth}
        \centering
        \includegraphics[width=\linewidth]{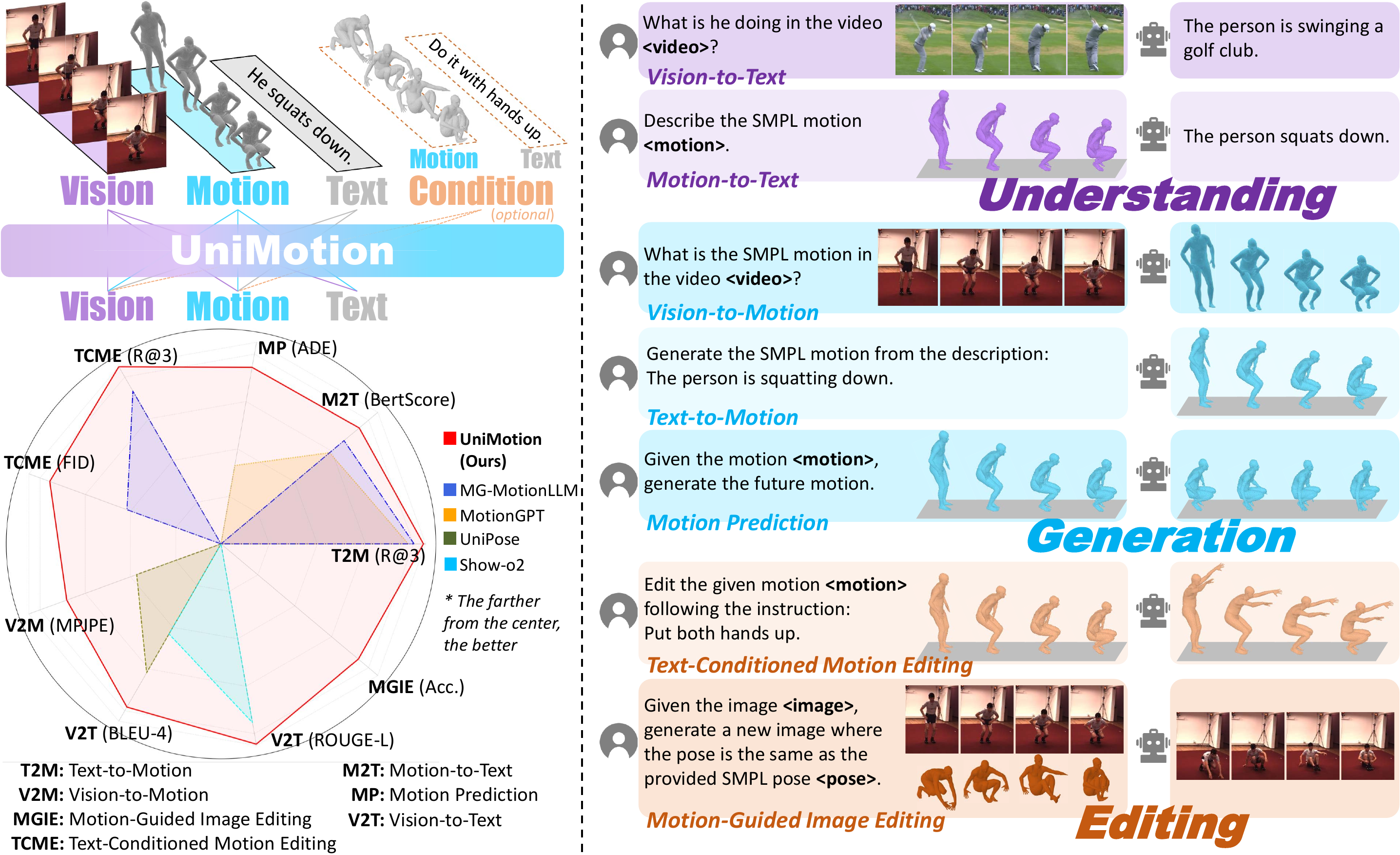}
        \vspace{-7mm}
        \captionof{figure}{\textbf{Left:} Overview and performance comparison of UniMotion, a unified framework for any-to-any Motion, Text, and Vision understanding, generation, and editing. UniMotion is the first model to support all seven tri-modal tasks and achieves consistent superiority over existing methods. \textbf{Right:} Representative task demonstrations.}
        \label{fig:pipeline}
        \vspace{-8mm}
    \end{minipage}
\end{center}

\begingroup
\renewcommand{\thefootnote}{\ensuremath{*}}
\footnotetext{Equal contribution.}
\renewcommand{\thefootnote}{\ensuremath{\dagger}}
\footnotetext{Corresponding author: \texttt{liumengyuan@pku.edu.cn}.}
\endgroup

\input{sec/0_abstract}

\input{sec/1_intro}
\input{sec/2_related_work}
\input{sec/3_method}
\input{sec/4_exp}

\input{sec/5_conclusion}

\section*{Acknowledgement}
This work was supported by National Natural Science Foundation of China (No. 62473007), Guangdong Outstanding Youth Fund (No. 2026B1515020015), Shenzhen Innovation in Science and Technology Foundation for The Excellent Youth Scholars (No. RCYX20231211090248064).

%
%

\bibliographystyle{splncs04}
\bibliography{main}

\newpage
\input{sec/X_suppl}

\end{document}

%% file: sec/0_abstract.tex
\begin{abstract}
We present UniMotion, to our knowledge the first unified framework for simultaneous understanding and generation of human motion, natural language, and RGB images within a single architecture. Existing unified models handle only restricted modality subsets (e.g., Motion–Text or static Pose–Image) and predominantly rely on discrete tokenization, which introduces quantization errors and disrupts temporal continuity. UniMotion overcomes both limitations through a core principle: treating motion as a first-class continuous modality on equal footing with RGB. A novel Cross-Modal Aligned Motion VAE (CMA-VAE) and symmetric dual-path embedders construct parallel continuous pathways for Motion and RGB within a shared LLM backbone. To inject visual-semantic priors into motion representations without requiring images at inference, we propose Dual-Posterior KL Alignment (DPA), which distills a vision-fused encoder's richer posterior into the motion-only encoder. To address the cold-start problem—where text supervision alone is too sparse to calibrate the newly introduced motion pathway—we further propose Latent Reconstruction Alignment (LRA), a self-supervised pre-training strategy that uses dense motion latents as unambiguous conditions to co-calibrate the embedder, backbone, and flow head, establishing a stable motion-aware foundation for all downstream tasks. UniMotion achieves state-of-the-art performance across seven tasks spanning any-to-any understanding, generation, and editing among the three modalities, with especially strong advantages on cross-modal compositional tasks.
\keywords{Motion Generation \and Unified Framework \and MLLMs}
\end{abstract}

%% file: sec/1_intro.tex
\vspace{-9mm}
\section{Introduction}
\vspace{-3mm}

Recent advances in unified multimodal large language models (MLLMs) have demonstrated strong and general capabilities in jointly understanding and generating text and images~\cite{showo,showo2,janus,transfusion,wu2024next,team2024chameleon}, exhibiting strong cross-modal reasoning within a shared semantic space. However, human motion---a critical dynamic modality---has not been systematically integrated into such unified frameworks. Motion sequences encode rich temporal dynamics and spatial structure indispensable for game animation, embodied intelligence, virtual reality, medical rehabilitation, and privacy-preserving action analysis. Constructing a single framework that unifies the Motion-Text-RGB tri-modality while supporting both understanding and generation remains an unresolved key problem.

Existing approaches have not truly addressed this challenge. One line of work, exemplified by MotionGPT~\cite{motiongpt}, unifies Motion and Text through VQ-VAE tokenization~\cite{vqvae}, but lacks the ability to perceive or generate visual information. Another line, represented by UniPose~\cite{unipose}, integrates human \emph{pose} (single-frame body configuration) with vision and language, but is restricted to static estimation and image understanding, with no generation capability. Both rely on discrete tokenization, which inevitably introduces quantization errors\cite{vqvae} that disrupt temporal continuity and structural fidelity; moreover, the asymmetry between discrete tokens and the continuous RGB feature space complicates cross-modal alignment. In summary, prior methods handle only partial modality subsets, and none forms a unified generation-understanding system spanning continuous motion sequences, natural language, and real images.

To this end, we propose \textbf{UniMotion}: the first multimodal framework for unified understanding and generation across the Motion-Text-RGB tri-modality. As summarized in Fig.~\ref{fig:pipeline}, UniMotion covers understanding, generation, and editing across the three modalities within one model, together with a unified performance comparison against prior partial solutions. The core design philosophy is to treat motion as a continuous modality on equal footing with RGB and construct symmetric continuous pathways for both. Unlike discrete-tokenization methods, we represent motion with a continuous \textbf{C}ross-\textbf{M}odal \textbf{A}ligned Motion VAE (\textbf{CMA-VAE}) and, through a dual-path embedder, provide complementary semantic-awareness and fine-grained generation channels that enable natural cross-modal alignment inside the shared LLM backbone. Within the backbone, we further employ lightweight modality-routed LoRA---routing each token to a modality-specific low-rank adapter so that motion and text/RGB modalities can adapt the shared parameters independently---and hybrid attention that reconciles motion's need for bidirectional temporal interaction with text's autoregressive constraints, alongside modality-specific flow heads that predict velocity fields in the respective latent spaces. This symmetric continuous design eliminates quantization errors at the architecture level, naturally unifying the understanding and generation of motion and images.

Beyond the architecture, we propose two key innovations for cross-modal semantic alignment under heterogeneous supervision. First, we design \textbf{D}ual-\textbf{P}osterior KL \textbf{A}lignment (\textbf{DPA}) for CMA-VAE by jointly training a Vision-Fused Motion Encoder $q_\psi(z \mid \text{motion}, \text{image})$ and a Motion Encoder $q_\phi(z \mid \text{motion})$, minimizing the KL divergence between their posteriors. This allows the Motion Encoder to absorb image-provided semantic supervision during training while requiring only motion at inference. Datasets lacking paired images participate by omitting the DPA loss. Second, we identify a supervision mismatch: the motion the model must generate is dense and kinematically rich, yet the text it learns from is much sparser---capturing only coarse action semantics while omitting details such as stride length, limb coordination, and subtle temporal transitions. Training the generation pathway solely from such under-specified signals leads to ambiguous learning, instability, and degraded fidelity. Yet the CMA-VAE latent $z$---the model's own continuous motion encoding---already preserves the full kinematic structure faithfully in compact form. This raises a natural question: before learning from sparse text, can the model first learn to generate motion from its own most informative encoding? We answer this with \textbf{L}atent \textbf{R}econstruction \textbf{A}lignment (\textbf{LRA}), a simple yet effective self-supervised pre-training strategy that treats $z$ embeddings as ``dense motion prompts'' and trains the model to reconstruct $z$ from noise in latent space. This self-reconstruction provides precise, unambiguous geometric supervision that jointly calibrates the embedder, LLM backbone, and flow head, establishing a robust motion-aware pathway as the shared foundation for all downstream tasks.

Thanks to these designs, UniMotion achieves state-of-the-art results across virtually all downstream tasks, supporting true any-to-any understanding and generation among Motion, Text, and RGB. It demonstrates especially prominent advantages on cross-modal compositional tasks. Our main contributions are:

\vspace{-1mm}
\begin{enumerate}
    \item We propose UniMotion, the first framework to unify Motion, Text, and RGB understanding and generation in a single architecture, overcoming the modality-coverage and task-direction limitations of prior methods. Beyond reusing a pretrained Text--RGB MLLM, the key challenge we solve is inserting a kinematically-constrained \emph{continuous} motion modality---absent from such backbones---into the shared space.
    \item We propose a fully continuous motion paradigm: CMA-VAE encodes motion into a visual-semantically enriched continuous latent space, and symmetric dual-path embedders with hybrid attention and modality-routed LoRA construct parallel continuous pathways for Motion and RGB, eliminating quantization bottlenecks at the architecture level.
    \item We propose Dual-Posterior KL Alignment (DPA) and Latent Reconstruction Alignment (LRA) as complementary alignment strategies: DPA injects visual-semantic supervision into the motion encoder via posterior alignment, while LRA leverages dense motion latents for self-supervised pathway calibration---jointly constructing a well-aligned tri-modal space that improves training stability and performance across all tasks.
\end{enumerate}

%% file: sec/2_related_work.tex
\begin{table}[t]
\centering
\setlength\tabcolsep{1.2mm}
\renewcommand\arraystretch{1.1}
\caption{Comparison of UniMotion with representative methods from a task perspective. \cmark\ indicates the method supports the task; \xmark\ indicates it does not. UniMotion uniquely unifies comprehensive understanding, generation, and editing across Motion, Text, and RGB modalities within a single continuous motion-aware MLLM. ``Size'' is the backbone parameter count; UniMotion (1.5B) is smaller than the 7B MLLM baselines yet covers all seven tasks.}
\vspace{-3mm}
\label{tab:related_comparison}
\resizebox{\linewidth}{!}{%
\begin{tabular}{@{}lcccccccccc@{}}
\toprule
\multirow{2}{*}{Method} &
\multirow{2}{*}{Venue} &
\multirow{2}{*}{Size} &
\multicolumn{2}{c}{\textbf{Understanding}} &
\multicolumn{3}{c}{\textbf{Generation}} &
\multicolumn{2}{c}{\textbf{Editing}} &
\multirow{2}{*}{Motion Repr.} \\
\cmidrule(lr){4-5} \cmidrule(lr){6-8} \cmidrule(lr){9-10}
& & & M2T & V2T & T2M & V2M & Pred & Mot. Edit & MGIE & \\
\midrule
MotionGPT~\cite{motiongpt}     & NeurIPS'23 & 220M & \cmark\ & \xmark\ & \cmark\ & \xmark\ & \cmark\ & \xmark\ & \xmark\ & VQ-VAE (discrete)              \\
MG-MotionLLM~\cite{mgmotionllm}& CVPR'25    & 220M & \cmark\ & \xmark\ & \cmark\ & \xmark\ & \xmark\ & \cmark\ & \xmark\ & VQ-VAE (discrete)              \\
UniPose~\cite{unipose}         & CVPR'25    & 7B & \xmark\ & \cmark\ & \xmark\ & \cmark\ & \xmark\ & \xmark\ & \xmark\ & VQ-VAE pose tokens (discrete)  \\
HMVLM~\cite{hmvlm}             & NeurIPS'25 & 7B & \xmark\ & \cmark\ & \cmark\ & \cmark\ & \xmark\ & \xmark\ & \xmark\ & VQ-VAE body-part (discrete)    \\
Show-o2~\cite{showo2}          & NeurIPS'25   & 1.5/7B & \xmark\ & \cmark\ & \xmark\ & \xmark\ & \xmark\ & \xmark\ & \xmark\ & ---                            \\
\midrule
\rowcolor{mygray}
\textbf{UniMotion (Ours)}      & ---     & \textbf{1.5B} & \cmark\ & \cmark\ & \cmark\ & \cmark\ & \cmark\ & \cmark\ & \cmark\ & \textbf{CMA-VAE (continuous)} \\
\bottomrule
\end{tabular}%
}
\vspace{-4mm}
\end{table}

\section{Related Work}

Table~\ref{tab:related_comparison} summarizes the key distinctions between UniMotion and representative prior methods. We now discuss each related area below.

\vspace{-1mm}
\subsection{Human Motion Generation and Understanding}
\vspace{-1mm}

Human motion modeling covers text-driven generation (Text-to-Motion) and semantic understanding (Motion-to-Text), evolving from task-specific models to unified frameworks.

\noindent\textbf{Discrete tokenization paradigm.}
Discretizing motion into codebook indices via VQ-VAE~\cite{vqvae} has become mainstream. MotionGPT~\cite{motiongpt} treated motion as a ``foreign language'' for unified generation and understanding; MG-MotionLLM~\cite{mgmotionllm} advanced multi-granularity motion modeling. However, discretization inevitably introduces quantization errors, causing temporal jitter and detail loss, while codebook collapse limits diversity.

\noindent\textbf{Continuous representation paradigm.}
To overcome quantization limitations, another line models motion in continuous latent space~\cite{mld,zhang2024motiondiffuse}. MLD~\cite{mld} performs diffusion in the VAE\cite{kingma2013auto} latent space, balancing quality and efficiency. Recent approaches combine motion VAEs with diffusion-based heads for smoother, more realistic generation while preserving semantic control.

Despite progress on Motion-Text tasks (Table~\ref{tab:related_comparison}), these methods remain confined to the Motion-Text subspace, lacking visual perception and generation needed for Vision-to-Motion or motion-guided editing.

\vspace{-1mm}
\subsection{Vision-Pose Multimodal Large Language Models}
\vspace{-1mm}

With the rise of MLLMs such as LLaVA~\cite{llava}, researchers have begun integrating human pose and skeletons into these models to enhance fine-grained behavior understanding. UniPose~\cite{unipose} discretizes 3D pose into tokens via VQ-VAE for unified pose understanding and generation. HMVLM~\cite{hmvlm} introduces MoE LoRA-based instruction tuning, yet still relies on discrete body-part tokenization and pairwise modality connections rather than a unified latent space.

These methods share common limitations: (1) most handle only static pose or pairwise connections without modeling continuous motion; (2) discrete quantization introduces precision bottlenecks; (3) visual interaction is typically unidirectional, lacking image generation capability.

\vspace{-1mm}
\subsection{Unified Multimodal Understanding and Generation}
\vspace{-1mm}

Unifying understanding and generation of arbitrary modalities within a single architecture is a fundamental direction toward building general artificial intelligence. Show-o~\cite{showo} pioneered a single Transformer fusing autoregressive and discrete diffusion for joint text-image understanding and generation. Janus-Pro~\cite{januspro} substantially improved performance through data and model scale expansion; Show-o2~\cite{showo2} further advanced cross-modal capabilities, with recent work exploring personalization~\cite{li2026personalize} and efficient diffusion decoding~\cite{yuan2026visual}.

While these works have achieved significant advances in the Text-RGB domain, human motion---encoding human intent and behavioral logic---remains absent from the core of unified frameworks. Treating motion as video pixels is inefficient and ignores skeletal topology and kinematic constraints.

%% file: sec/3_method.tex
\section{Method}

\begin{figure}[tb]
    \centering
    \includegraphics[width=\linewidth]{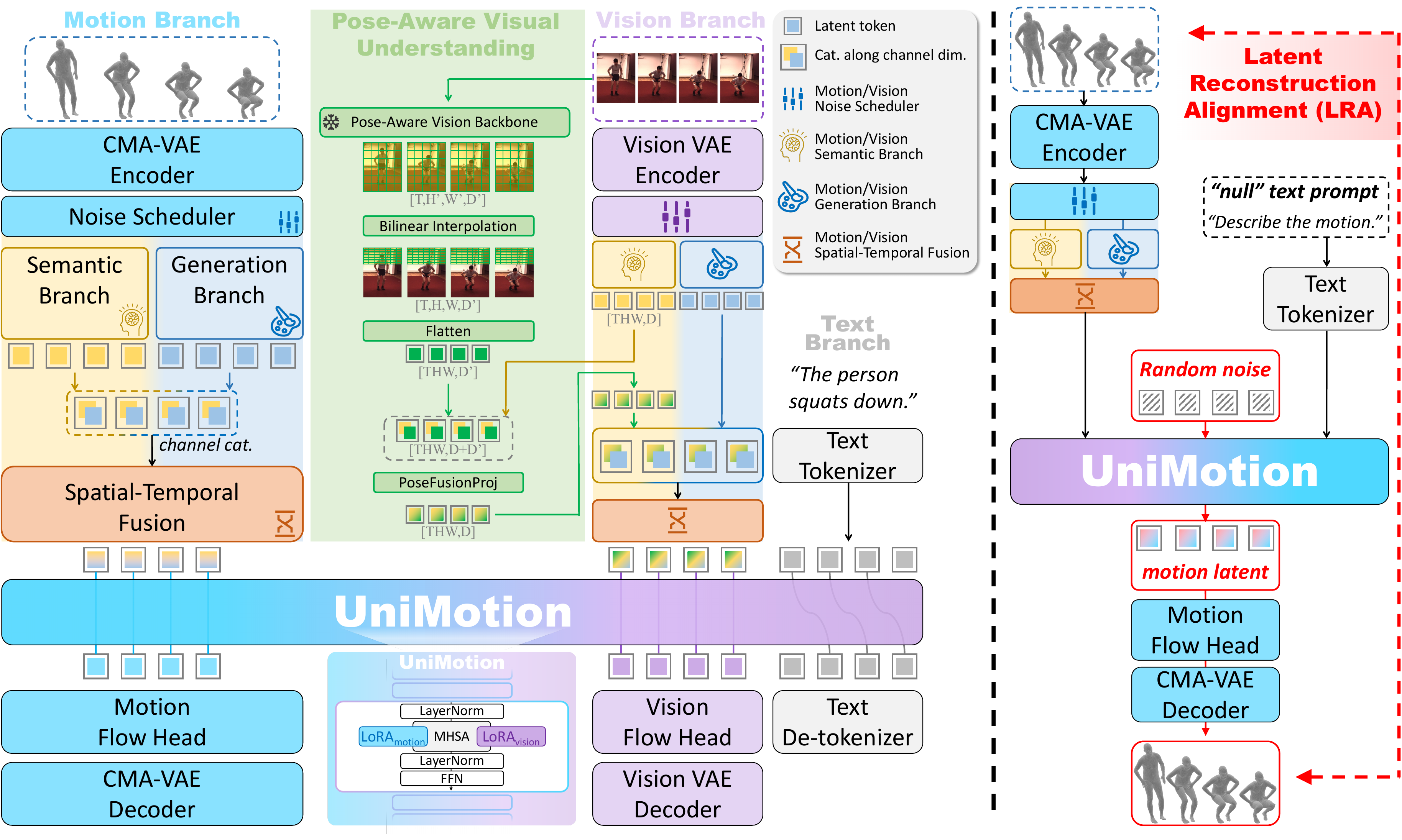}
    \caption{Overview of UniMotion. \textbf{(Left)}~ UniMotion unifies motion, text, and RGB through symmetric continuous pathways: motion and images are encoded into continuous latents (via CMA-VAE and a vision VAE), mapped by a dual-path embedder that separates semantic abstraction from detail-preserving generation, and processed by a shared backbone for both multimodal understanding and modality-specific flow-based synthesis. \textbf{(Right)}~ Latent Reconstruction Alignment (LRA) pre-trains the motion pathway with a self-supervised Motion-to-Motion task, using motion latents as dense, unambiguous conditions to reconstruct motion from noise, thereby co-calibrating the embedder, backbone, and motion head before all downstream tri-modal learning.}
    \label{fig:method_pipeline}
    \vspace{-5mm}
\end{figure}

\vspace{-1mm}
\subsection{Overview}
\label{sec:overview}
\vspace{-1mm}

UniMotion aims to unify modeling of Motion, Text, and RGB within a single framework, simultaneously supporting understanding and generation tasks (including T2M, M2T, Motion Prediction, Motion Editing, Vision-to-Motion, Vision-to-Text, and Motion-guided Image Editing (MGIE)). The core design philosophy is to treat motion as a continuous modality on equal footing with images and construct symmetric continuous pathways for both.

Unlike prior methods that rely on discrete VQ-VAE tokenization for representing motion dynamics (\eg, MotionGPT, UniPose), we adopt continuous VAE representations for motion, which offers two key advantages: (1) it avoids irreversible information loss during quantization, preserving the temporal continuity and structural fidelity of motion; (2) continuous motion latents share a symmetric representational form with the continuous RGB pathway, enabling cross-modal alignment naturally at the architecture level.

The overall framework, illustrated in Fig.~\ref{fig:method_pipeline}, comprises three core components: (1) \textbf{CMA-VAE} (Sec.~\ref{sec:cmavae}): encodes motion sequences into continuous latent representations with implicit visual-semantic supervision injected via Dual-Posterior KL Alignment (DPA); (2) \textbf{Unified Multimodal Architecture} (Sec.~\ref{sec:arch}): built upon Show-o2~\cite{showo2}, with symmetric dual-path embedders for Motion and RGB, a shared LLM backbone, and modality-specific flow\cite{flowmatching} heads; (3) \textbf{Latent Reconstruction Alignment (LRA)} (Sec.~\ref{sec:lic}): using dense motion prompts to first warm up the motion pathway (the embedder, flow head, and motion-routed adaptation), followed by progressive multi-stage fine-tuning.

\vspace{-1mm}
\subsection{Cross-Modal Aligned Motion VAE (CMA-VAE)}
\label{sec:cmavae}
\vspace{-1mm}

\begin{figure}[t]
    \centering
    \includegraphics[width=0.78\linewidth]{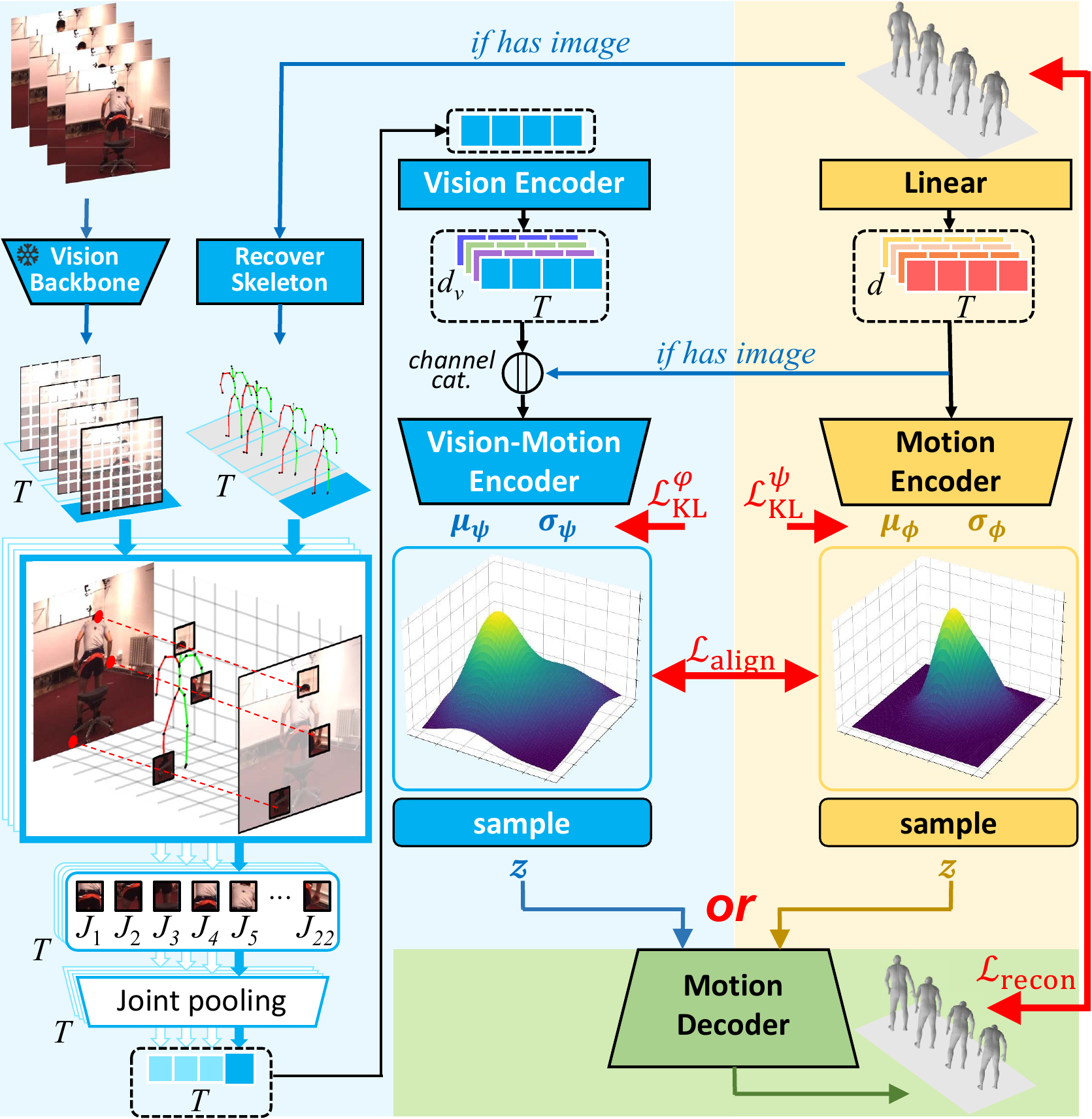}
    \vspace{-2mm}
    \caption{CMA-VAE with DPA. CMA-VAE learns a continuous motion latent space using a motion-only encoder for inference and a vision-fused encoder for training-time visual supervision. When paired images are available, motion-guided visual features are fused with motion and distilled via DPA, enabling the shared decoder to learn visually informed motion latents without requiring images at inference.}
    \label{fig:cmavae}
    \vspace{-4mm}
\end{figure}

CMA-VAE encodes variable-length motion sequences into continuous low-dimensional latent representations while injecting implicit visual-semantic supervision through Dual-Posterior KL Alignment (DPA). As shown in Fig.~\ref{fig:cmavae}, its core consists of three components: Motion Encoder $q_\phi(z \mid \mathbf{m})$, Vision-Fused Motion Encoder $q_\psi(z \mid \mathbf{m}, \mathbf{v})$, and a shared Motion Decoder $p_\xi(\mathbf{m} \mid z)$. Unlike standard motion VAEs~\cite{mld} that encode motion in isolation, CMA-VAE is cross-modal by construction: its motion-only inference encoder is coupled to a vision-fused training-time encoder via posterior alignment, so the latent space carries visual-semantic structure.

\noindent\textbf{Motion Encoder.}
The Motion Encoder is used at inference. Given a motion sequence $\mathbf{m} \in \mathbb{R}^{T \times D_m}$, a linear layer maps each frame to the latent space, followed by learnable positional encodings and a SkipTransformer Encoder. A linear head predicts Gaussian parameters $(\mu_\phi, \log\sigma^2_\phi)$, and the latent code is obtained via reparameterization $z = \mu_\phi + \sigma_\phi \odot \epsilon$, $\epsilon \sim \mathcal{N}(0, I)$, with $z \in \mathbb{R}^{T_z \times d}$.

\noindent\textbf{Vision-Fused Motion Encoder.}
Used only during training, this encoder fuses motion and visual information. Its motion branch shares the front-end with the Motion Encoder, producing $\mathbf{h}_\mathrm{motion} \in \mathbb{R}^{T_z \times d}$. The vision branch extracts spatial features from RGB image $\mathbf{v}$ via a frozen HRNet\cite{sun2019deep}, then applies bilinear grid sampling at motion-guided 2D joint positions $\mathbf{j}_{2d}(\mathbf{m})$---skeleton projections derived from the motion sequence:

\vspace{-2mm}
\begin{equation}
    \mathbf{f}_\mathrm{vis} = \mathrm{VisionEnc}\bigl(\mathrm{GridSample}(\mathrm{HRNet}(\mathbf{v}),\, \mathbf{j}_{2d}(\mathbf{m}))\bigr) \in \mathbb{R}^{T_z \times d_v}.
\end{equation}
\vspace{-2mm}

After joint-dimension pooling, motion and visual features are concatenated and processed by an independent SkipTransformer Encoder to yield $(\mu_\psi, \log\sigma^2_\psi)$ and sample $z_\mathrm{fused}$.

\noindent\textbf{Motion Decoder.}
The decoder accepts latent $z$, passes it through positional encoding and SkipTransformer layers, and maps back to the $D_m$-dimensional motion space via a linear layer.

\vspace{-1mm}
\subsubsection{Dual-Posterior KL Alignment (DPA)}

The core idea of DPA is: constrain the Motion Encoder posterior $q_\phi(z \mid \mathbf{m})$ to approximate the Vision-Fused posterior $q_\psi(z \mid \mathbf{m}, \mathbf{v})$, so that the Motion Encoder implicitly absorbs visual-semantic supervision during training while requiring only motion input at inference. The total CMA-VAE training objective is:

\vspace{-1mm}
\begin{equation}
    \mathcal{L}_\mathrm{VAE} = \mathcal{L}_\mathrm{recon} + \lambda_\mathrm{KL}\,\bigl(\mathcal{L}_\mathrm{KL}^{\phi} + \mathcal{L}_\mathrm{KL}^{\psi}\bigr) + \lambda_\mathrm{align} \cdot \mathcal{L}_\mathrm{align}.
\end{equation}
\vspace{-2mm}

\noindent\textbf{Reconstruction loss.}
For samples with paired images, $z_\mathrm{fused}$ from the Vision-Fused Encoder is used for decoding; otherwise $z_\mathrm{motion}$ from the Motion Encoder is used:

\vspace{-3mm}
\begin{equation}
    \mathcal{L}_\mathrm{recon} = \frac{1}{|\mathcal{M}|}\sum_{i \in \mathcal{M}} \mathrm{SmoothL1}(\hat{m}_i, m_i).
\end{equation}
\vspace{-1mm}

\noindent\textbf{KL regularization.}
Both encoder posteriors are independently regularized toward the standard normal prior via the standard Gaussian KL $\mathcal{L}_\mathrm{KL}^{\star} = \frac{1}{2}\sum_{k=1}^{d}(\mu_{\star,k}^2 + \sigma_{\star,k}^2 - \log\sigma_{\star,k}^2 - 1)$, $\star \in \{\phi,\, \psi\}$.

\noindent\textbf{DPA alignment loss.}
Computed only for samples with paired images, this term distills visual-semantic knowledge into the Motion Encoder:

\vspace{-2mm}
\begin{equation}
    \mathcal{L}_\mathrm{align} = D_\mathrm{KL}\bigl(q_\phi(z \mid \mathbf{m}) \,\|\, q_\psi(z \mid \mathbf{m}, \mathbf{v})\bigr).
\end{equation}
\vspace{-2mm}

\noindent For two diagonal Gaussians $\mathcal{N}(\mu_\phi, \sigma^2_\phi)$ and $\mathcal{N}(\mu_\psi, \sigma^2_\psi)$, the closed-form KL is:

\vspace{-1mm}
\begin{equation}
    D_\mathrm{KL} = \frac{1}{2}\sum_{k=1}^{d}\left(\log\frac{\sigma_{\psi,k}^2}{\sigma_{\phi,k}^2} + \frac{\sigma_{\phi,k}^2 + (\mu_{\phi,k} - \mu_{\psi,k})^2}{\sigma_{\psi,k}^2} - 1\right).
\end{equation}
\vspace{-1mm}

$\mathcal{L}_\mathrm{align}$ uses a linear warm-up schedule\cite{goyal2017accurate} to prevent strong alignment from disrupting reconstruction in early training.
We adopt $D_\mathrm{KL}(q_\phi \| q_\psi)$ with the Vision-Fused posterior $q_\psi$ detached as the alignment target. In the knowledge-distillation sense (student$\to$teacher), this is the reverse KL, which is mode-seeking: it drives $q_\phi$ to concentrate on the most salient semantic modes of $q_\psi$, yielding compact, high-confidence motion representations that capture the core visual-semantic information while filtering out view-specific visual noise. The forward direction $D_\mathrm{KL}(q_\psi \| q_\phi)$ would be mode-covering, forcing $q_\phi$ to spread over all modes of $q_\psi$---including noisy or irrelevant visual modes---leading to an over-dispersed posterior that dilutes representational precision.

\noindent\textbf{Flexible data utilization.}
For datasets without paired images (\eg, HumanML3D), $\mathcal{L}_\mathrm{align}$ is simply dropped. For datasets with paired images (\eg, Human3.6M), all three losses apply jointly. At inference, the Vision-Fused Encoder is entirely discarded, ensuring no overhead.

\vspace{-2mm}
\subsection{Unified Multimodal Architecture}
\label{sec:arch}
\vspace{-1mm}

UniMotion is built upon Show-o2~\cite{showo2}, extending it with a Motion modality pathway architecturally symmetric to RGB.

\vspace{-1mm}
\subsubsection{Dual-Path Embedder}

Given the CMA-VAE latent $z \in \mathbb{R}^{T_z \times d}$, two parallel branches process $z$: a \textbf{Semantic branch} (MLP $+$ Transformer Encoder layers) extracts high-level semantic features, mirroring SigLIP\cite{zhai2023sigmoid} on the vision side; a \textbf{Generation branch} (MLP $+$ learnable positional encodings) maps $z$ directly to the LLM hidden dimension, preserving fine-grained motion details and mirroring the vision PatchEmbed. The two outputs are concatenated and projected to the unified LLM hidden dimension via RMSNorm $+$ MLP. All tasks uniformly use the fused embeddings. This dual-path design is particularly advantageous for motion-conditioned synthesis tasks (\eg, Motion Editing and Prediction), where the two branches structurally decouple semantic comprehension from fine-grained detail preservation.
To further strengthen visual-motion alignment, the RGB pathway is additionally equipped with a pose-aware vision backbone: initialized from a pretrained human body encoder~\cite{tokenhmr} and kept frozen, it extracts body-structure-aware features that complement SigLIP's global visual semantics. Both streams are fused into the same RGB token representation via projection, maintaining architectural symmetry with the Motion embedder.

\subsubsection{Hybrid Attention and Modality-Routed LoRA}

\textbf{Hybrid attention} maintains global causal ordering at the sequence level while enabling bidirectional full attention within each motion span, reconciling the flow matching objective---which requires simultaneous velocity field prediction across the entire motion latent---with text autoregressive generation. Modality-routed LoRA assigns separate low-rank adaptation branches for Text/RGB and Motion tokens in each attention layer, enabling modality-specific adaptation with only $\sim$2\% additional parameters while preserving the LLM's existing capabilities.

\vspace{-1mm}
\subsubsection{Modality-Specific Flow Heads}

For generation tasks, LLM hidden states are transformed back to the target modality latent space via modality-specific flow heads.

\noindent\textbf{Motion flow head.}
A lightweight AdaLN-conditioned structure (Modulated Attention Blocks + MotionFinalLayer) maps backbone features to velocity predictions $\hat{v}_m \in \mathbb{R}^{T_z \times d_m}$ in the motion latent space. Timestep conditioning is injected via timestep embedding $\mathbf{c}_t$; the output layer is zero-initialized to stabilize early training.

\noindent\textbf{Vision flow head.}
An isomorphic design whose output dimension aligns with the vision latent representation. This ``shared backbone + modality-specific head'' design balances parameter efficiency and cross-modal adaptation.

\vspace{-1mm}
\subsection{Latent Reconstruction Alignment (LRA)}
\label{sec:lic}
\vspace{-1mm}

\noindent\textbf{The cold-start problem and dense self-supervision.}
After DPA pre-training, the motion pathway is still uncalibrated: the embedder, motion flow head, and motion-routed adaptation have not yet been jointly aligned. Direct multi-task training from this state causes clear degradation (T2M R@3 only 0.801 vs.\ our final 0.841, see Sec.~\ref{sec:ablation_lic}). While text descriptions are inherently sparse (one-to-many mapping), the CMA-VAE latent $z \in \mathbb{R}^{T_z \times d}$ is a dense, lossless encoding whose self-reconstruction constitutes an unambiguous one-to-one mapping---an ideal zero-cost pre-training signal for bootstrapping the motion pathway.

\noindent\textbf{M2M task design.}
We instantiate this via a Motion-to-Motion (M2M) self-reconstruction task. The CMA-VAE Encoder produces $z$; the dual-path embedder projects $z$ into the LLM, whose hidden states pass through the motion flow head to reconstruct $z$ from noise:

\vspace{-1mm}
\begin{equation}
    \mathcal{L}_\mathrm{M2M} = \mathbb{E}_{z_0 \sim \mathcal{N}(0,I),\, t \sim p(t)}\bigl[\| v_\theta(z_t,\, t \mid \mathrm{Embed}_\mathrm{fused}(z)) - (z - z_0) \|^2\bigr],
\end{equation}
\vspace{-1mm}

where $z_t = t \cdot z + (1-t) \cdot z_0$. Critically, the LLM receives $\mathrm{Embed}_\mathrm{fused}(z)$ as conditioning rather than the noised $z_t$ (injected only into the flow head via AdaLN), ensuring the motion pathway learns to structurally encode motion semantics rather than merely acting as a denoiser.

\noindent\textbf{Co-calibration and cross-task transfer.}
M2M simultaneously co-calibrates the embedder (compressing $z$ into LLM-readable tokens), the motion-routed adaptation in the shared backbone (extracting structural cues), and the flow head (mastering latent-space geometry) in a tightly coupled manner---a calibration that sparse text supervision alone cannot provide due to ambiguous geometric feedback. Once calibrated, this pathway becomes the shared foundation for all downstream tasks: T2M benefits from the pre-calibrated flow head, M2T reuses the embedder's semantic compression, and Vision$\to$M leverages the aligned pipeline so the LLM can focus purely on cross-modal mapping. Crucially, LRA does not degenerate into a trivial identity mapping: the dual-path embedder compresses $z$ into LLM-compatible tokens via non-trivial projection and Transformer encoding, and the flow head must predict velocity fields from Gaussian noise conditioned on LLM hidden states---a probabilistic mapping that is architecturally incapable of bypassing the LLM.

%% file: sec/4_exp.tex
\vspace{-1mm}
\section{Experiments}
\vspace{-1mm}

Our experiments span three dimensions: (1) \textbf{Unification}---systematic comparison with SOTA on tasks spanning Motion, Text, and RGB (Sec.~\ref{sec:sota}); (2) \textbf{Continuity}---ablation validating the superiority of CMA-VAE (Sec.~\ref{sec:abl_cmavae}); (3) \textbf{Alignment}---ablation confirming the key contributions of DPA and LRA (Sec.~\ref{sec:abl_dpa}).

\vspace{-3mm}
\subsection{Experimental Setup}
\vspace{-1mm}
\label{sec:exp_setup}

\noindent\textbf{Datasets.} We evaluate UniMotion on HumanML3D~\cite{humanml3d} (Text-to-Motion, Motion-to-Text, Motion Prediction), MotionFix~\cite{motionfix} (Motion Editing), Human3.6M~\cite{h36m} (Vision-to-Motion, Vision-to-Text), and MoVid~\cite{chen2024motionllm} (Vision-to-Text). We also construct a triplet evaluation set from Human3.6M and 3DPW for Motion-guided Image Editing (MGIE). We adopt a unified 269-dimensional motion representation to support both generation and body recovery tasks.

\begin{table}[t]
\centering
\setlength\tabcolsep{1.5mm}
\renewcommand\arraystretch{1.0}
\caption{Unified multi-task comparison using one representative metric per task. ``N/A'' indicates the method does not support the task. UniMotion is the only method covering all seven tasks.}
\vspace{-2mm}
\label{tab:unified}
\resizebox{\linewidth}{!}{%
\begin{tabular}{@{}lccccccc@{}}
\toprule
\textbf{Method} & \textbf{\shortstack{T2M \\ R@3$\uparrow$}} & \textbf{\shortstack{M2T \\ BertScore$\uparrow$}} & \textbf{\shortstack{MotionPred \\ ADE$\downarrow$}} & \textbf{\shortstack{MotionEdit \\ R@3$\uparrow$}} & \textbf{\shortstack{V2M \\ MPJPE$\downarrow$}} & \textbf{\shortstack{V2T \\ BLEU-4$\uparrow$}} & \textbf{\shortstack{MGIE \\ Mot.Acc$\uparrow$}} \\
\midrule
MotionGPT~\cite{motiongpt}    & 0.778 & 32.4 & 4.745 & \cellcolor[gray]{0.9}N/A & \cellcolor[gray]{0.9}N/A & \cellcolor[gray]{0.9}N/A & \cellcolor[gray]{0.9}N/A \\
MG-MotionLLM~\cite{mgmotionllm} & 0.802 & 36.7 & \cellcolor[gray]{0.9}N/A & 73.23 & \cellcolor[gray]{0.9}N/A & \cellcolor[gray]{0.9}N/A & \cellcolor[gray]{0.9}N/A \\
UniPose~\cite{unipose}        & \cellcolor[gray]{0.9}N/A & \cellcolor[gray]{0.9}N/A & \cellcolor[gray]{0.9}N/A & \cellcolor[gray]{0.9}N/A & 81.8 & 17.3 & \cellcolor[gray]{0.9}N/A \\
Show-o2~\cite{showo2}         & \cellcolor[gray]{0.9}N/A & \cellcolor[gray]{0.9}N/A & \cellcolor[gray]{0.9}N/A & \cellcolor[gray]{0.9}N/A & \cellcolor[gray]{0.9}N/A & 12.1 & \cellcolor[gray]{0.9}N/A \\
\midrule
\rowcolor{mygray}
\textbf{UniMotion (Ours)} & \textbf{0.841} & \textbf{41.2} & \textbf{3.172} & \textbf{84.94} & \textbf{75.0} & \textbf{21.9} & \textbf{0.67} \\
\bottomrule
\vspace{-4mm}
\end{tabular}%
}
\end{table}

\noindent\textbf{Implementation details.}
The unified backbone is based on Show-o2 1.5B~\cite{showo2}, equipped with a dual-path motion embedder and a motion flow head. Modality-routed LoRA (rank 32) is applied to all attention layers. We train with AdamW and use Euler ODE for generation inference.
All training is conducted on 4$\times$A6000 GPUs.

\vspace{-3mm}
\subsection{Comparison with State-of-the-Art}
\label{sec:sota}
\vspace{-1mm}

\begin{table}[t]
\centering
\setlength\tabcolsep{1.5mm}
\renewcommand\arraystretch{1.0}
\caption{Text-to-Motion generation on HumanML3D. $\rightarrow$ indicates closer to \textit{Real} is better. UniMotion$^\dagger$ is the single-task model; UniMotion is the unified multi-task model. \textbf{Bold}: best; \underline{underline}: second best.}
\vspace{-2mm}
\label{tab:t2m}
\resizebox{\linewidth}{!}{%
\begin{tabular}{@{}llcccccc@{}}
\toprule
\textbf{Type} & \textbf{Method} & \textbf{R@1$\uparrow$} & \textbf{R@2$\uparrow$} & \textbf{R@3$\uparrow$} & \textbf{FID$\downarrow$} & \textbf{MMDist$\downarrow$} & \textbf{DIV$\rightarrow$} \\
\midrule
--- & Real & 0.511 & 0.703 & 0.797 & 0.002 & 2.974 & 9.503 \\
\midrule
\multirow{4}{*}{Gen.\ only}
& T2M-GPT~\cite{t2mgpt}         & 0.491 & 0.680 & 0.775 & 0.116 & 3.118 & 9.761 \\
& DiverseMotion~\cite{diversemotion} & 0.515 & 0.706 & 0.802 & \underline{0.072} & 2.941 & 9.683 \\
& MoMask~\cite{momask}           & 0.521 & 0.713 & 0.807 & \textbf{0.045} & 2.958 & 9.620 \\
& \textbf{UniMotion$^\dagger$}   & \underline{0.528} & \underline{0.726} & \underline{0.830} & 0.231 & \underline{2.845} & 9.649 \\
\midrule
\multirow{5}{*}{Gen.\ \& Und.}
& TM2T~\cite{tm2t}               & 0.424 & 0.618 & 0.729 & 1.501 & 3.467 & 8.589 \\
& MotionGPT~\cite{motiongpt}     & 0.492 & 0.681 & 0.778 & 0.232 & 3.096 & \textbf{9.528} \\
& HMVLM~\cite{hmvlm}             & 0.463 & 0.646 & 0.744 & 0.156 & 3.328 & \underline{9.544} \\
& MG-MotionLLM~\cite{mgmotionllm} & 0.516 & 0.706 & 0.802 & 0.303 & 2.952 & 9.960 \\
\rowcolor{mygray}
& \textbf{UniMotion}             & \textbf{0.557} & \textbf{0.749} & \textbf{0.841} & 0.194 & \textbf{2.715} & 9.583 \\
\bottomrule
\vspace{-5mm}
\end{tabular}%
}
\end{table}

\begin{table}[t]
\centering
\setlength\tabcolsep{1.5mm}
\renewcommand\arraystretch{1.0}
\vspace{-1mm}
\caption{Motion-to-Text understanding on HumanML3D. Evaluation protocol follows~\cite{tm2t}. UniMotion$^\dagger$ is the single-task model. \textbf{Bold}: best; \underline{underline}: second best.}
\vspace{-2mm}
\label{tab:m2t}
\resizebox{\linewidth}{!}{%
\begin{tabular}{@{}lcccccccc@{}}
\toprule
\textbf{Method} & \textbf{R@1$\uparrow$} & \textbf{R@3$\uparrow$} & \textbf{MMDist$\downarrow$} & \textbf{Bleu@1$\uparrow$} & \textbf{Bleu@4$\uparrow$} & \textbf{Rouge$\uparrow$} & \textbf{CIDEr$\uparrow$} & \textbf{BertScore$\uparrow$} \\
\midrule
Real           & 0.523 & 0.828 & 2.901 & --- & --- & --- & --- & --- \\
\midrule
TM2T~\cite{tm2t}              & 0.516 & 0.823 & 2.935 & 48.9 & 7.00 & 38.1 & 16.8 & 32.2 \\
MotionGPT~\cite{motiongpt}    & 0.543 & 0.827 & 2.821 & 48.2 & 12.47 & 37.4 & 29.2 & 32.4 \\
LaMPM2T~\cite{lamp}           & 0.547 & 0.831 & 2.808 & 47.8 & 13.04 & 37.1 & 28.9 & 32.7 \\
MG-MotionLLM~\cite{mgmotionllm} & \textbf{0.592} & \textbf{0.866} & 2.581 & --- & 8.06 & --- & --- & 36.7 \\
\midrule
\textbf{UniMotion$^\dagger$}           & 0.547 & 0.843 & \underline{2.562} & \underline{56.9} & \underline{18.6} & \underline{44.7} & \underline{35.4} & \underline{38.5} \\
\rowcolor{mygray}
\textbf{UniMotion}            & \underline{0.562} & \underline{0.857} & \textbf{2.481} & \textbf{60.4} & \textbf{20.7} & \textbf{46.5} & \textbf{39.3} & \textbf{41.2} \\
\bottomrule
\end{tabular}%
}
\vspace{-3mm}
\end{table}

\subsubsection{Unified Multi-task Comparison}

Table~\ref{tab:unified} provides a unified comparison across all seven tasks using one representative metric per task for transparent cross-task evaluation. UniMotion is the only method covering all tasks.

\vspace{-2mm}
\subsubsection{Text-to-Motion Generation}

\begin{figure}[t]
    \centering
    \includegraphics[width=\linewidth]{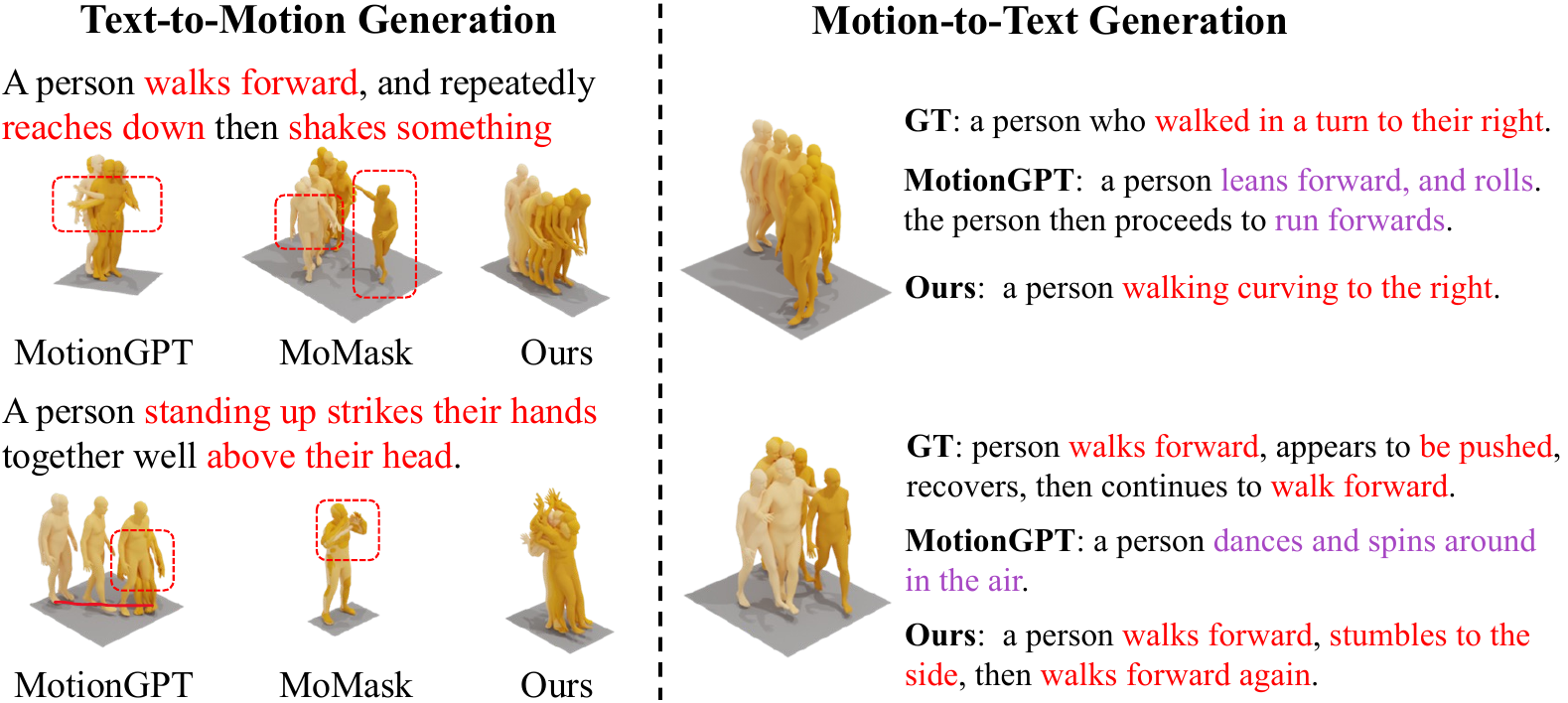}
    \caption{Qualitative comparison on T2M and M2T.}
    \label{fig:qual_t2m}
    \vspace{-4mm}
\end{figure}

Table~\ref{tab:t2m} compares UniMotion with SOTA on HumanML3D. UniMotion$^\dagger$ denotes the task-specific variant sharing the same architecture and training recipe, trained only on the evaluated task without cross-task supervision; UniMotion is the unified multi-task model.

UniMotion achieves the best R-Precision and MMDist by clear margins across compared methods, establishing strong semantic alignment between text and generated motion. Notably: (1) the substantial gains on these alignment metrics reflect CMA-VAE's continuous representation and DPA's effectiveness in binding motion-text semantics; (2) the unified multi-task model consistently outperforms the single-task variant (UniMotion$^\dagger$), demonstrating positive cross-modal transfer from RGB supervision; (3) while single-task discrete methods obtain lower FID (MoMask: 0.045)---reflecting their advantage in single-distribution fitting---UniMotion's leading semantic alignment highlights the cross-modal reasoning enabled by continuous representations within a unified framework. Furthermore, Diversity (9.583) closely matches the real data (9.503), confirming that alignment improvements preserve sample variability. As qualitatively confirmed in Fig.~\ref{fig:qual_t2m}, MoMask's low FID does not prevent it from missing critical spatial constraints (\eg, hands only reaching shoulder level instead of above the head), while UniMotion faithfully renders fine-grained modifiers and produces more accurate M2T captions than MotionGPT.

\vspace{-1mm}
\subsubsection{Motion-to-Text Understanding}
As shown in Table~\ref{tab:m2t}, UniMotion leads all caption-quality metrics by substantial margins (Bert-Score: 41.2 vs.\ 36.7; CIDEr: 39.3 vs.\ 29.2; Bleu@4: 20.7 vs.\ 13.04), indicating semantically faithful and detail-rich descriptions rather than generic templates. We attribute this to CMA-VAE's continuous latents preserving kinematic nuances (\eg, stride patterns, joint articulations) that discrete tokenization discards. MG-MotionLLM achieves higher retrieval R@k, reflecting stronger embedding-level matching, while UniMotion's advantage lies in generation quality. The unified model further surpasses the single-task baseline (BertScore: 41.2 vs.\ 38.5), confirming positive cross-modal transfer from DPA's visual-semantic supervision.

\subsubsection{Motion Prediction and Text-Conditioned Motion Editing}
UniMotion outperforms all baselines on both tasks (Table~\ref{tab:pred_edit}). Beyond CMA-VAE's continuous latents for preserving temporal dynamics, DPA and visual co-training inject visual-geometry priors (\eg, plausible body configurations, global balance) that are especially helpful for forecasting future poses and editing local joints while maintaining whole-body coherence. For motion prediction, consistent gains in FID (0.871 vs.\ 0.905), ADE (3.172 vs.\ 4.745), and FDE (5.068 vs.\ 6.040) reflect these advantages. For motion editing, UniMotion substantially surpasses MG-MotionLLM (R@1: 63.81 vs.\ 47.96; R@3: 84.94 vs.\ 73.23; FID: 0.170 vs.\ 0.409), confirming that continuous representation combined with visual grounding enables precise text-guided modification at the joint level. Architecturally, the dual-path embedder is critical for these tasks: its semantic--generative decoupling enables the source motion to simultaneously provide high-level structural conditioning and fine-grained kinematic detail for the target.
\vspace{0.5mm}

\begin{table}[t]
\caption{(a) Motion prediction on AMASS (HumanML3D subset), following MotionGPT~\cite{motiongpt}. (b) Text-conditioned motion editing on MotionFix; R@k is generated-to-target retrieval precision (\%).}
\vspace{-2mm}
\label{tab:pred_edit}
\centering
\footnotesize
\setlength{\tabcolsep}{3pt}
\renewcommand\arraystretch{1.0}
\begin{minipage}[t]{0.5\linewidth}
\centering
\textbf{(a)} Motion Prediction\\[2pt]
\begin{tabular}{@{}lccc@{}}
\toprule
\textbf{Method} & \textbf{FID$\downarrow$} & \textbf{ADE$\downarrow$} & \textbf{FDE$\downarrow$} \\
\midrule
Real & 0.002 & --- & --- \\
MDM~\cite{mdm} & 6.031 & 5.446 & 8.561 \\
MotionGPT~\cite{motiongpt} & 0.905 & 4.745 & 6.040 \\
\rowcolor{mygray}
\textbf{UniMotion} & \textbf{0.871} & \textbf{3.172} & \textbf{5.068} \\
\bottomrule
\end{tabular}
\end{minipage}%
\begin{minipage}[t]{0.5\linewidth}
\centering
\textbf{(b)} Motion Editing\\[2pt]
\begin{tabular}{@{}lccc@{}}
\toprule
\textbf{Method} & \textbf{R@1$\uparrow$} & \textbf{R@3$\uparrow$} & \textbf{FID$\downarrow$} \\
\midrule
GT & 100.0 & 100.0 & --- \\
MDM~\cite{mdm} & 39.10 & 54.84 & 0.917 \\
MG-MotionLLM~\cite{mgmotionllm} & 47.96 & 73.23  & 0.409 \\
\rowcolor{mygray}
\textbf{UniMotion} & \textbf{63.81} & \textbf{84.94} & \textbf{0.170} \\
\bottomrule
\end{tabular}
\vspace{-3mm}
\end{minipage}
\vspace{-2mm}
\end{table}

\begin{table}[t]
\centering
\setlength\tabcolsep{1.5mm}
\renewcommand\arraystretch{1.0}
\caption{Comparison of different motion representation methods. APE and AVE denote \emph{mean-pose} Absolute Position Error and Average Velocity Error, respectively, in cm and cm/s. All variants were trained independently during the pure reconstruction phase, followed by multi-task fine-tuning under the same LLM backbone. \textbf{Bold}: best; \underline{underline}: second best.}
\vspace{-2mm}
\label{tab:abl_repr}
\resizebox{\linewidth}{!}{%
\begin{tabular}{@{}lccccccc@{}}
\toprule
\multirow{2}{*}{\textbf{Motion Repr.}} & \multicolumn{3}{c}{\textbf{Reconstruction Quality}} & \multirow{2}{*}{\textbf{\shortstack[c]{T2M\\[-0.5pt]R@3$\uparrow$}}} & \multirow{2}{*}{\textbf{\shortstack[c]{M2T\\[-0.5pt]BertScore$\uparrow$}}} & \multirow{2}{*}{\textbf{\shortstack[c]{Pred\\[-0.5pt]ADE$\downarrow$}}} & \multirow{2}{*}{\textbf{\shortstack[c]{Edit\\[-0.5pt]R@3$\uparrow$}}} \\
\cmidrule(lr){2-4}
& \textbf{APE$\downarrow$} & \textbf{AVE$\downarrow$} & \textbf{FID$\downarrow$} & & & & \\
\midrule
VQ-VAE (MotionGPT\cite{motiongpt})    & 17.15 & \underline{0.813} & \underline{0.0674} & 0.771 & 35.0 & 4.713 & 69.71 \\
MLD-VAE (MLD\cite{mld})         & \underline{9.28} & 0.981 & 0.1283 & \underline{0.810} & \underline{37.2} & \underline{3.784} & \underline{77.46} \\
\rowcolor{mygray}
\textbf{CMA-VAE (Ours)} & \textbf{3.53} & \textbf{0.428} & \textbf{0.0282} & \textbf{0.841} & \textbf{41.2} & \textbf{3.172} & \textbf{84.94} \\
\bottomrule
\end{tabular}%
}
\vspace{-3mm}
\end{table}

\begin{table*}[t]
\centering
\small
\begin{minipage}[t][0.33\textheight][t]{0.48\textwidth}
\centering
\renewcommand\arraystretch{1.1}
\caption{Vision-to-Motion on Human3.6M (H3.6M). \textbf{Bold}: best in category for each group and evaluation setting under the same protocol.}
\vspace{-2mm}
\label{tab:rgb2m}
\resizebox{\linewidth}{!}{%
\begin{tabular}{@{}lccc@{}}
\toprule
\textbf{Method} & \textbf{Type} & \textbf{MPJPE$\downarrow$} & \textbf{PA-MPJPE$\downarrow$} \\
\midrule
\multicolumn{4}{c}{\textit{Specialist Models}} \\\midrule
HMR~\cite{hmr} & Spec. & 100.7 & 67.7 \\
PyMAF~\cite{pymaf2021} & Spec. & 64.2 & 44.9 \\
SMPLer~\cite{smpler} & Spec. & \textbf{50.8} & 37.3 \\
HMR2.0~\cite{hmr2} & Spec. & 52.2 & 37.1 \\
TokenHMR~\cite{tokenhmr} & Spec. & 52.4 & 36.8 \\
Zolly~\cite{wang2023zolly} & Spec. & 55.0 & \textbf{35.9} \\
\midrule
\multicolumn{4}{c}{\textit{MLLM-based Models}} \\\midrule
ChatPose~\cite{chatpose} & MLLM & 146.7 & 92.4 \\
UniPose~\cite{unipose} & MLLM & 81.8 & 50.9 \\
\rowcolor{mygray}
\textbf{UniMotion} & MLLM & \textbf{75.0} & \textbf{46.1} \\
\bottomrule
\end{tabular}}
\vfill
\end{minipage}\hfill
\begin{minipage}[t][0.30\textheight][t]{0.48\textwidth}
\centering
\renewcommand\arraystretch{1.1}
\caption{Motion-guided Image Editing (MGIE). UP+So2 is a two-stage text-mediated pipeline; OP+CN renders SMPL-derived skeletons as spatial conditions. Mot.\ Acc.\ is the hit rate based on HMR2.0-estimated pose, using PA-MPJPE $\leq 100.0$\,mm as success.}
\vspace{-3mm}
\label{tab:mgie}
\resizebox{\linewidth}{!}{%
\begin{tabular}{@{}lccc@{}}
\toprule
\textbf{Method} & \textbf{FID$\downarrow$} & \textbf{CLIP$\uparrow$} & \textbf{Mot.Acc$\uparrow$} \\
\midrule
UniPose+Show-o2~\cite{unipose,showo2} & 26.16 & 0.22 & 0.50 \\
OpenPose+ControlNet~\cite{openpose,controlnet} & 22.34 & 0.29 & 0.59 \\
\rowcolor{mygray}
\textbf{UniMotion} & \textbf{18.92} & \textbf{0.31} & \textbf{0.67} \\
\bottomrule
\end{tabular}}
\vspace{2mm}
\caption{Ablation of DPA and LRA on T2M, M2T, Motion Prediction, Motion Editing, and Vision-to-Motion.}
\vspace{-3mm}
\label{tab:abl_dpa}
\label{tab:abl_lic}
\resizebox{\linewidth}{!}{%
\begin{tabular}{@{}lccccc@{}}
\toprule
\textbf{Setting} & \textbf{\shortstack{T2M \\ R@3$\uparrow$}} & \textbf{\shortstack{M2T \\ BertScore$\uparrow$}} & \textbf{\shortstack{MotionPred \\ ADE$\downarrow$}} & \textbf{\shortstack{MotionEdit \\ R@3$\uparrow$}} & \textbf{\shortstack{V2M \\ MPJPE$\downarrow$}} \\
\midrule
w/o DPA & 0.818 & 38.4 & 3.654 & 80.35  & 83.1 \\
w/o LRA & 0.801 & 38.1 & 3.777 & 78.72 & 84.3 \\
\rowcolor{mygray}
\textbf{Full UniMotion} & \textbf{0.841} & \textbf{41.2} & \textbf{3.172} & \textbf{84.94} & \textbf{75.0} \\
\bottomrule
\end{tabular}}
\vfill
\end{minipage}
\end{table*}

\subsubsection{Vision-to-Motion and Motion-guided Image Editing (MGIE)}

As reported in Table~\ref{tab:rgb2m}, among MLLM methods, UniMotion significantly surpasses the strong UniPose baseline on both MPJPE (75.0 vs.\ 81.8) and PA-MPJPE (46.1 vs.\ 50.9), benefiting from DPA's visual-geometry priors and CMA-VAE's precise continuous encoding. The remaining gap to specialist methods is expected for a general-purpose framework.
For V2M we follow UniPose's single-reference-image protocol; the input image is encoded through the RGB pathway at inference, while the CMA-VAE vision-fused encoder used for DPA is discarded after training. Performance is driven mainly by our DPA/LRA alignment rather than the frozen pose-aware backbone (which alone reaches only 86.9 vs.\ our 75.0 MPJPE), and the model transfers zero-shot to in-the-wild 3DPW (93.6 MPJPE), indicating robustness beyond clean indoor imagery.

For MGIE, UniMotion is the first unified motion-aware MLLM supporting end-to-end motion-conditioned image generation in a shared latent space, without explicit skeleton rendering or text mediation. Against UniPose+Show-o2~\cite{unipose,showo2} and OpenPose+ControlNet~\cite{openpose,controlnet}, it outperforms both on all metrics (Table~\ref{tab:mgie}), with the largest margin on Motion Accuracy (0.67 vs.\ 0.59), confirming that end-to-end latent-space reasoning avoids the information loss of staged pipelines.

\vspace{-3mm}
\subsection{Ablation Studies}
\label{sec:ablation}
\vspace{-1mm}

\subsubsection{CMA-VAE Motion Representation Comparison}
\label{sec:abl_cmavae}

To systematically evaluate CMA-VAE, we compare against VQ-VAE (MotionGPT~\cite{motiongpt}) and MLD-VAE (MLD~\cite{mld}) in Table~\ref{tab:abl_repr}. VQ-VAE suffers the worst absolute position accuracy (APE=17.15) due to irreversible quantization errors, and transfers weakest to downstream tasks (T2M R@3=0.771, Edit R@3=69.71); its seemingly lower AVE (0.813) and FID (0.0674) compared to MLD-VAE reflect the bounded codebook artificially limiting deviations rather than faithful reconstruction. MLD-VAE improves global positioning (APE: 9.28) yet exhibits a position--velocity trade-off (AVE: 0.981, FID: 0.1283), confirming that a plain continuous VAE without cross-modal anchoring loosens temporal dynamics. Nevertheless, MLD-VAE substantially outperforms VQ-VAE on all downstream tasks, demonstrating that continuous latent spaces inherently facilitate LLM cross-modal alignment regardless of absolute reconstruction quality.

CMA-VAE, calibrated by DPA, resolves this trade-off and achieves the best results on all metrics---both reconstruction (APE=3.53, AVE=0.428, FID=0.0282) and downstream transfer (T2M R@3=0.841, Edit R@3=84.94). Adding DPA to the same architecture improves all tasks (T2M R@3: 0.818$\to$0.841; Edit R@3: 80.35$\to$84.94), confirming that implicit visual-semantic supervision is the core driver of CMA-VAE's superiority.

\vspace{-1mm}
\subsubsection{DPA and LRA Ablation}
\label{sec:abl_dpa}
\label{sec:ablation_lic}

As shown in Table~\ref{tab:abl_dpa}, removing either component causes consistent performance drops across all tasks. Without LRA, the motion pathway is insufficiently calibrated by sparse cross-modal supervision alone, leading to drops on T2M (R@3: 0.801 vs.\ 0.841), M2T (BertScore: 38.1 vs.\ 41.2), Motion Prediction (ADE: 3.777 vs.\ 3.172), Motion Editing (R@3: 78.72 vs.\ 84.94), and Vision$\to$M (MPJPE: 84.3 vs.\ 75.0). Without DPA, the representation loses explicit visual-semantic alignment and likewise remains consistently below the full model across all tasks. Notably, DPA's gains extend to HumanML3D-only tasks (T2M, M2T) despite lacking paired images in that dataset, evidencing implicit knowledge transfer through shared encoder parameters. The two mechanisms are complementary: LRA calibrates the motion pathway's geometric capacity via dense self-supervision, while DPA enriches representational semantics through cross-modal posterior alignment.

%% file: sec/5_conclusion.tex
\vspace{-2mm}
\section{Conclusion}
\vspace{-2mm}

We presented UniMotion, to our knowledge the first continuous Motion-Text-RGB unified framework in the motion-LLM setting for human-centric multimodal reasoning and unified understanding and generation within a single architecture. By treating motion as a continuous modality on equal footing with RGB, UniMotion constructs symmetric continuous pathways through CMA-VAE and a dual-path embedder, avoiding the quantization artifacts of discrete tokenization. Dual-Posterior KL Alignment and Latent Reconstruction Alignment jointly establish a well-aligned tri-modal latent space, yielding strong results across diverse tasks with clear advantages on cross-modal compositional tasks. As current tri-modal benchmarks are predominantly single-person, our evaluation focuses on single-person motion; the per-span hybrid attention is compatible with multi-person interaction, which we leave as future work.

%% file: sec/X_suppl.tex
\setcounter{section}{0}
\renewcommand{\thesection}{\Alph{section}}

In this supplementary material, we provide comprehensive additional results, detailed architectural specifications, and in-depth analyses to further validate the effectiveness and reproducibility of our proposed framework, \textbf{UniMotion}. The content is organized as follows:

\begin{itemize}
    \item \textbf{Sec. \ref{sec:suppl_results} (Additional Experimental Results)} presents extended quantitative results for Vision-to-Text and exhaustive ablation studies on the Dual-Path embedder, motion representations (CMA-VAE vs. VQ), hybrid attention, and modality-routed LoRA.
    
    \item \textbf{Sec. \ref{sec:suppl_qual} (Qualitative Results)} provides extensive visualizations across all seven tasks—including text-to-motion, motion-to-text, prediction, editing, and vision-conditioned synthesis—alongside a multi-domain (spatial, frequency, and temporal) analysis of motion reconstruction quality.
    
    \item \textbf{Sec. \ref{sec:suppl_analysis} (Further Analysis)} details the mathematical intuition behind DPA (mode-seeking distillation), proves the non-triviality of LRA via information bottleneck analysis, and evaluates the model's zero-shot generalization to out-of-distribution datasets like 3DPW.
    
    \item \textbf{Sec. \ref{sec:suppl_arch} (Architecture and Implementation Details)} specifies the 269-dimensional unified motion representation, the CMA-VAE design (including motion-guided sampling), and the formulation of our flow matching generation heads and pose-aware vision backbone.
    
    \item \textbf{Sec. \ref{sec:suppl_train} (Training Pipeline, Data, and Evaluation)} outlines the progressive multi-stage training configuration, details our tri-modal data construction strategies (H3.6M, MotionFix, etc.), and clarifies the evaluation protocols for all benchmark metrics.
    
    \item \textbf{Sec. \ref{sec:suppl_limit} (Limitations and Broader Impact)} discusses the current constraints regarding computational overhead and domain-specific visual alignment, while reflecting on the potential societal benefits and ethical considerations of motion-aware AI.
\end{itemize}

\section{Additional Experimental Results}
\label{sec:suppl_results}

\subsection{Vision-to-Text}
\vspace{-1mm}

\begin{table}[htbp]
\centering
\small
\renewcommand\arraystretch{1.1}
\caption{Vision-to-Text on H3.6M. \textbf{Bold}: best; \underline{underline}: second best.}
\vspace{-3mm}
\label{tab:rgb2t}
\resizebox{0.7\linewidth}{!}{%
\begin{tabular}{@{}lcccc@{}}
\toprule
\textbf{Method} & \textbf{Param.} & \textbf{BLEU-4$\uparrow$} & \textbf{ROUGE-L$\uparrow$} & \textbf{METEOR$\uparrow$} \\
\midrule
Show-o2~\cite{showo2} & 1.5B   & 12.1    & 33.9  & 35.7  \\
Qwen-2.5-VL~\cite{qwenvl}    & 7B    & 16.5 & \underline{37.4} & \underline{39.6} \\
UniPose~\cite{unipose}     & 7B  & \underline{17.3} & 34.9  & 38.6 \\
\rowcolor{mygray}
\textbf{UniMotion (Ours)} & 1.5B & \textbf{21.9} & \textbf{38.0} & \textbf{41.7} \\
\bottomrule
\end{tabular}%
}
\vspace{-2mm}
\end{table}

We first report \textbf{Vision-to-Text} on H3.6M, which probes whether the visual pathway can extract human pose semantics from visual observations and translate them into natural language. UniMotion substantially surpasses both general-purpose MLLMs and specialized UniPose on all metrics with a significantly smaller parameter count (1.5B vs.\ 7B). The improvement over UniPose (BLEU-4: 21.9 vs.\ 17.3) is attributed to the CMA-VAE motion encoder learning fine-grained visual-semantic priors via DPA training, combined with the pose-aware vision backbone providing body-structure-aware features.

We then extend the same Vision-to-Text interface from the H3.6M setting to temporally richer video inputs. UniMotion uniformly samples frames from each video and processes them through the image pathway. We evaluate this on the MoVid dataset~\cite{chen2024motionllm}, which contains diverse real-world human motion videos with descriptive captions. The corresponding qualitative visualizations are presented later in Sec.~\ref{sec:suppl_rgb_grounded_qual}, where we separately show H3.6M Vision-to-Text examples (single-frame) and MoVid Vision-to-Text examples (multi-frame) to distinguish the static and dynamic settings.

\subsection{Architecture Design Validation}
\label{sec:abl_arch}

We ablate two orthogonal design dimensions in the motion processing pipeline.
\textbf{(1) Embedder design}: \emph{Gen-Branch Only} retains only the Generation Branch (MLP direct projection $+$ learnable positional encoding, mirroring PatchEmbed on the RGB side, preserving fine-grained kinematic details at the cost of semantic abstraction); \emph{Sem-Branch Only} retains only the Semantic Branch (MLP $+$ $N_s{=}4$ Transformer Encoder layers, mirroring SigLIP, providing high-level semantic features at the cost of detail preservation); \emph{Dual-Path} combines both via RMSNorm$+$MLP fusion (our full design, Sec.~3.3 of main paper).
\textbf{(2) Motion representation}: \emph{VQ} uses MotionGPT-style discrete tokenization ($K{=}512$ codebook); \emph{VAE (w/o DPA)} is a plain continuous VAE sharing CMA-VAE's architecture but trained without the DPA alignment loss; \emph{CMA-VAE} is our full Cross-Modal Aligned Motion VAE trained with DPA.
VQ representations are architecturally incompatible with DPA's continuous Gaussian posterior alignment; comparisons are therefore within equivalent conditions per representation class.

\begin{table}[htbp]
\centering
\renewcommand\arraystretch{1.1}
\caption{Architecture design ablation using representative metrics. \emph{Gen-Branch Only}: Generation Branch only (MLP $+$ PosEmbed). \emph{Sem-Branch Only}: Semantic Branch only (MLP $+$ Transformer Encoder). \emph{Dual-Path}: both branches fused (ours). \emph{VAE (w/o DPA)}: plain continuous VAE, same architecture as CMA-VAE but without DPA training. \emph{CMA-VAE}: with DPA. VQ is incompatible with DPA by design. All models share the same hyperparameters and training duration.}
\vspace{-1mm}
\label{tab:abl_arch}
\resizebox{\linewidth}{!}{%
\begin{tabular}{@{}llcccc@{}}
\toprule
\textbf{Embedder Design} & \textbf{Repr.} & \textbf{\shortstack{T2M\\R@3$\uparrow$}} & \textbf{\shortstack{M2T\\BertScore$\uparrow$}} & \textbf{\shortstack{Edit\\R@3$\uparrow$}} & \textbf{\shortstack{Pred\\ADE$\downarrow$}} \\
\midrule
Gen-Branch Only  & VQ-VAE              & 0.752 & 34.2 & 64.57  & 5.128  \\
Gen-Branch Only  & CMA-VAE         & 0.824 & 36.3 & 70.40  & 4.862  \\
Sem-Branch Only  & CMA-VAE         & 0.798 & 40.1 & 68.85  &  4.181 \\
\midrule
Dual-Path        & VQ-VAE              & 0.771 & 35.0 & 69.71 & 4.713 \\
Dual-Path        & MLD-VAE (MLD\cite{mld})   & 0.810 & 37.2 & 77.46 & 3.784 \\
Dual-Path        & VAE (w/o DPA)       & 0.818 & 38.4 & 80.35 & 3.654 \\
\rowcolor{mygray}
\textbf{Dual-Path (Ours)} & \textbf{CMA-VAE} & \textbf{0.841} & \textbf{41.2} & \textbf{84.94} & \textbf{3.172} \\
\bottomrule
\end{tabular}%
}
\vspace{-2mm}
\end{table}


Four observations emerge from Table~\ref{tab:abl_arch}.
\textbf{(1) Continuous representations consistently outperform VQ across all embedder designs.}
Replacing VQ with CMA-VAE yields substantial gains at every level of path complexity, confirming that quantization errors fundamentally constrain both generation fidelity and cross-modal alignment regardless of the embedder architecture.
\textbf{(2) The two branches exhibit complementary functional specialization.}
Sem-Branch Only achieves higher M2T BertScore (40.1 vs.\ 36.3 for Gen-Branch Only), reflecting the Transformer encoder's stronger semantic compression; Gen-Branch Only leads on T2M R@3 (0.824 vs.\ 0.798) and Motion Editing (70.40 vs.\ 68.85), where fine-grained kinematic detail must be preserved for the flow head.
Neither branch alone approaches Dual-Path performance, confirming functional complementarity rather than redundancy.
\textbf{(3) DPA provides targeted gains beyond VAE continuity.}
Dual-Path $+$ VAE (w/o DPA) already achieves T2M R@3 0.818 and Edit R@3 80.35; adding DPA pushes these to 0.841 and 84.94, with the largest incremental gains on tasks that benefit from injected visual-semantic priors.
\textbf{(4) Dual-Path is essential for motion-conditioned synthesis.}
The performance gap between single-branch and Dual-Path variants is largest for Editing and Prediction---tasks where motion serves simultaneously as conditioning input and generation target---because the semantic branch preserves high-level structural intent while the generation branch maintains fine-grained joint-level detail.
Gen-Branch Only $+$ CMA-VAE on Motion Prediction (ADE 4.862) underperforms MotionGPT (4.745), while Dual-Path $+$ CMA-VAE (3.172) substantially surpasses it, confirming that the semantic--generative decoupling of the dual-path design is architecturally essential.

\subsection{Hybrid Attention Ablation}

\begin{table}[htbp]
\vspace{-1em}
\centering
\renewcommand\arraystretch{1.1}
\caption{Hybrid attention ablation on T2M, M2T, and motion editing using representative metrics.}
\vspace{-1mm}
\label{tab:abl_attn}
\resizebox{\linewidth}{!}{%
\begin{tabular}{@{}lccc@{}}
\toprule
\textbf{Attention Strategy} & \textbf{\shortstack{T2M\\R@3$\uparrow$}} & \textbf{\shortstack{M2T\\BertScore$\uparrow$}} & \textbf{\shortstack{Edit\\R@3$\uparrow$}} \\
\midrule
Global Causal Only                              & 0.825 & 39.3 & 79.6 \\
\rowcolor{mygray}
\textbf{Hybrid: Causal + Intra-Motion Full (Ours)} & \textbf{0.841} & \textbf{41.2} & \textbf{84.94} \\
\bottomrule
\end{tabular}%
}
\vspace{-2mm}
\end{table}

Global causal attention restricts motion tokens to unidirectional temporal interaction, lowering T2M alignment (R@3: 0.825 vs.\ 0.841) and motion editing precision (R@3: 79.6 vs.\ 84.94). Hybrid attention reconciles the needs of motion generation and language modeling: full attention within motion spans matches the flow matching training objective, while global causal ordering preserves text autoregressive modeling and improves motion-to-text understanding (BertScore: 41.2 vs.\ 39.3).

\subsection{Modality-Routed LoRA Ablation}

\begin{table}[htbp]
\vspace{-1em}
\centering
\renewcommand\arraystretch{1.1}
\caption{Modality-routed LoRA ablation on four representative tasks.}
\vspace{-1mm}
\label{tab:abl_lora}
\resizebox{\linewidth}{!}{%
\begin{tabular}{@{}lcccc@{}}
\toprule
\textbf{LoRA Strategy} & \textbf{\shortstack{T2M\\R@3$\uparrow$}} & \textbf{\shortstack{M2T\\BertScore$\uparrow$}} & \textbf{\shortstack{V2M\\MPJPE$\downarrow$}} & \textbf{\shortstack{V2T\\BLEU-4$\uparrow$}} \\
\midrule
LLM Frozen (No LoRA)             & 0.752 & 34.1 & 99.6 & 13.7 \\
Shared LoRA                      & 0.818 & 38.8 & 90.4 & 18.8 \\
\rowcolor{mygray}
\textbf{Routed LoRA (Ours)}      & \textbf{0.841} & \textbf{41.2} & \textbf{75.0} & \textbf{21.9} \\
\bottomrule
\end{tabular}%
}
\vspace{-2mm}
\end{table}

Routed LoRA achieves the best results across all tasks. The \textit{LLM Frozen} baseline under-adapts to motion-centric supervision, while Shared LoRA couples heterogeneous gradients from different modalities within a single parameter branch. Routed LoRA decouples modality-specific adaptation with deterministic routing and only $\sim$2\% additional parameters.

\section{Qualitative Results}
\label{sec:suppl_qual}

\subsection{Text-to-Motion Generation}

\begin{figure}[htbp]
    \centering
    \includegraphics[width=0.90\linewidth]{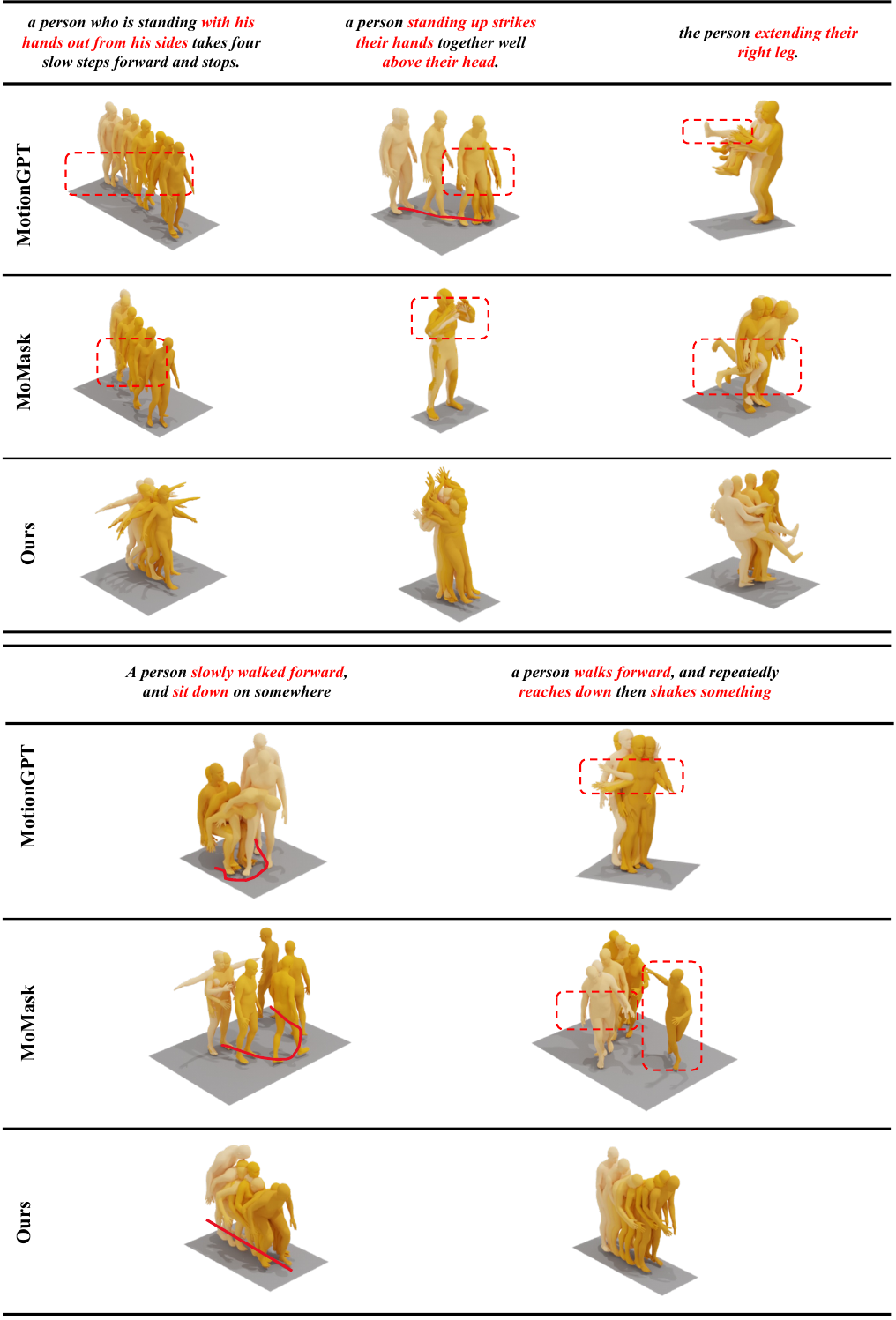}
    \vspace{-2mm}
    \caption{Qualitative comparison of text-driven motion generation on HumanML3D~\cite{humanml3d}. Each column corresponds to one text prompt, and rows show outputs from MotionGPT~\cite{motiongpt}, MoMask~\cite{momask}, and UniMotion. Red text highlights the key prompt constraints, while red dashed boxes mark prompt--motion mismatches in the baseline outputs, including missing body-part constraints, incorrect motion trajectories, and weak temporal modifiers. UniMotion produces motions with closer prompt correspondence and more coherent temporal transitions.}
    \label{fig:suppl_t2m}
    \vspace{-2mm}
\end{figure}

Figure~\ref{fig:suppl_t2m} highlights a consistent qualitative pattern across diverse prompts. In the first row, UniMotion better preserves explicit limb constraints such as \emph{``with his hands out''}, \emph{``hands together above their head''}, and \emph{``extending their right leg''}, whereas the baselines often under-execute the target arm or leg motion. In the lower examples, UniMotion also more faithfully captures temporal modifiers and action composition, such as \emph{walking forward then sitting} and \emph{walking while repeatedly reaching down}, indicating stronger control over both local pose details and global motion progression.

\subsection{Motion-to-Text Understanding}

\begin{figure}[htbp]
    \centering
    \includegraphics[width=\linewidth]{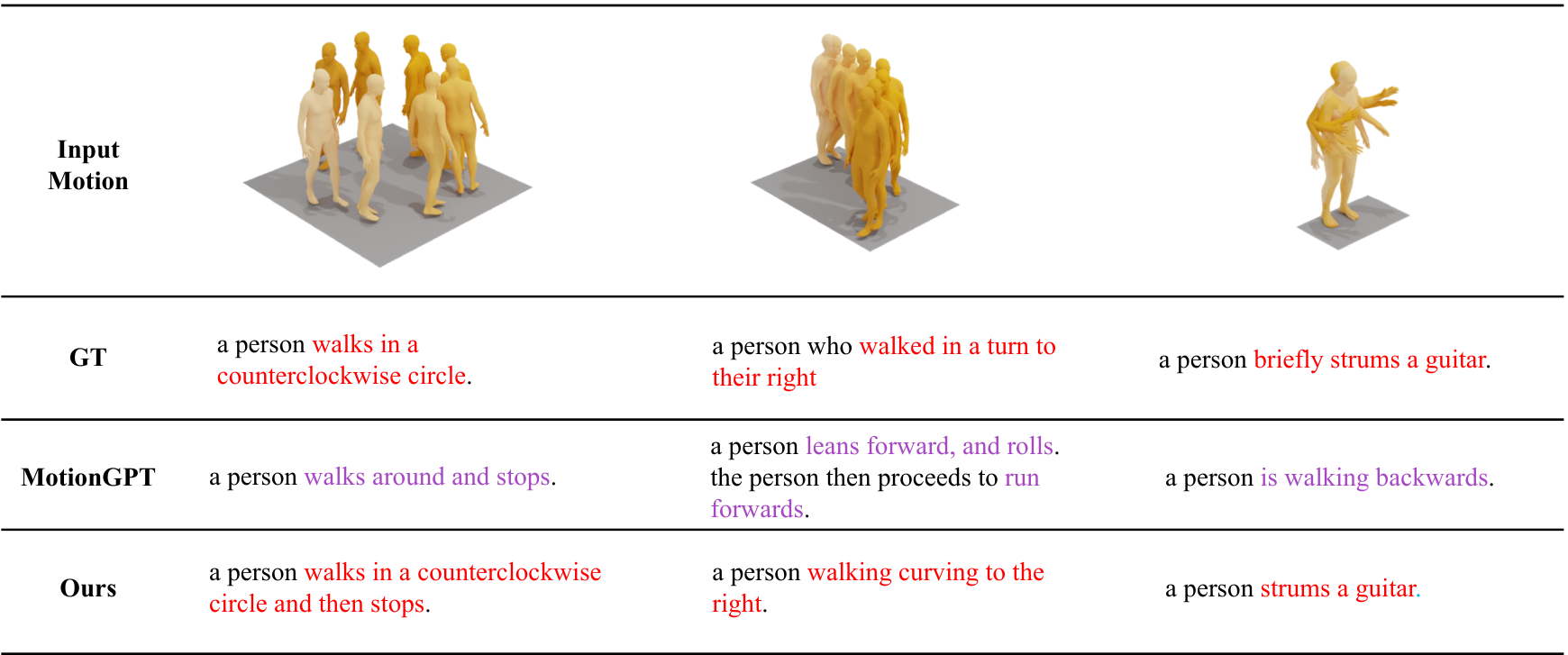}
    \vspace{-2mm}
    \caption{Qualitative comparison of motion captioning (Motion-to-Text). Each column shows the input motion, the ground-truth caption, the MotionGPT prediction, and the UniMotion prediction. Red phrases in the ground-truth and UniMotion captions highlight the key motion semantics, while purple phrases in the MotionGPT output indicate misaligned descriptions or hallucinated actions. UniMotion more precisely translates fine-grained joint articulations and temporal sequence states into accurate, fluent natural language.}
    \label{fig:suppl_m2t}
    \vspace{-2mm}
\end{figure}

Figure~\ref{fig:suppl_m2t} further illustrates that UniMotion tends to recover the dominant motion semantics while avoiding the generic or incorrect descriptions produced by MotionGPT. For example, UniMotion correctly identifies \emph{walking in a counterclockwise circle}, \emph{curving to the right}, and \emph{strumming a guitar}, whereas the baseline either drops the directional cue, introduces unrelated actions, or misclassifies the action entirely. This qualitative behavior is consistent with the quantitative gains in caption-quality metrics reported in the main paper.

\subsection{Motion Prediction}

\begin{figure}[htbp]
    \centering
    \includegraphics[width=\linewidth]{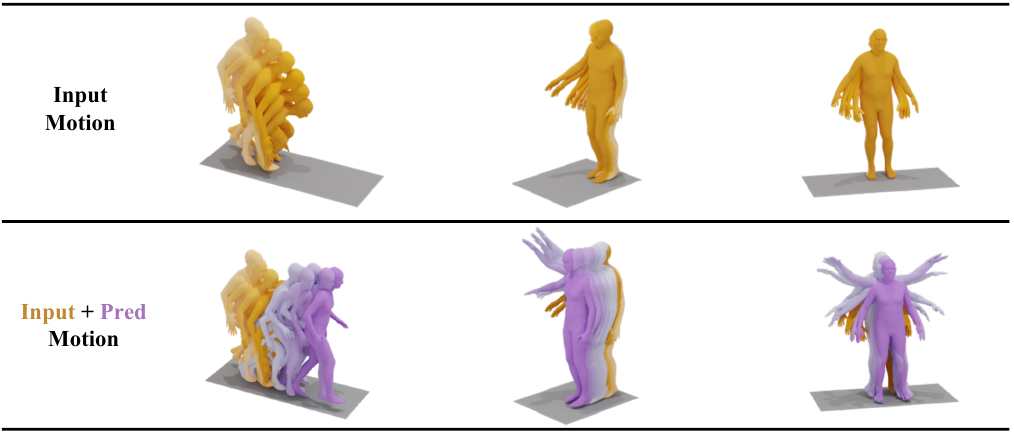}
    \vspace{-2mm}
    \caption{Qualitative visualization of motion prediction. Each column shows the observed input prefix (top) and the predicted continuation overlaid on the input motion (bottom), where the input segment is shown in yellow and the forecasted future is shown in purple. UniMotion extrapolates future motion with more consistent global trajectory direction, body balance, and temporally smooth joint evolution over long horizons.}
    \label{fig:suppl_pred}
    \vspace{-2mm}
\end{figure}

Figure~\ref{fig:suppl_pred} shows representative motion forecasting examples. Given the observed prefix, UniMotion extrapolates future body trajectories with stable global balance and temporally smooth limb evolution. Compared with discrete baselines, the predicted continuation better preserves action trend consistency, especially in turning, stepping, and arm-swing phases where accumulated drift is most visible.

\subsection{Text-Conditioned Motion Editing}

\begin{figure}[htbp]
    \centering
    \includegraphics[width=\linewidth]{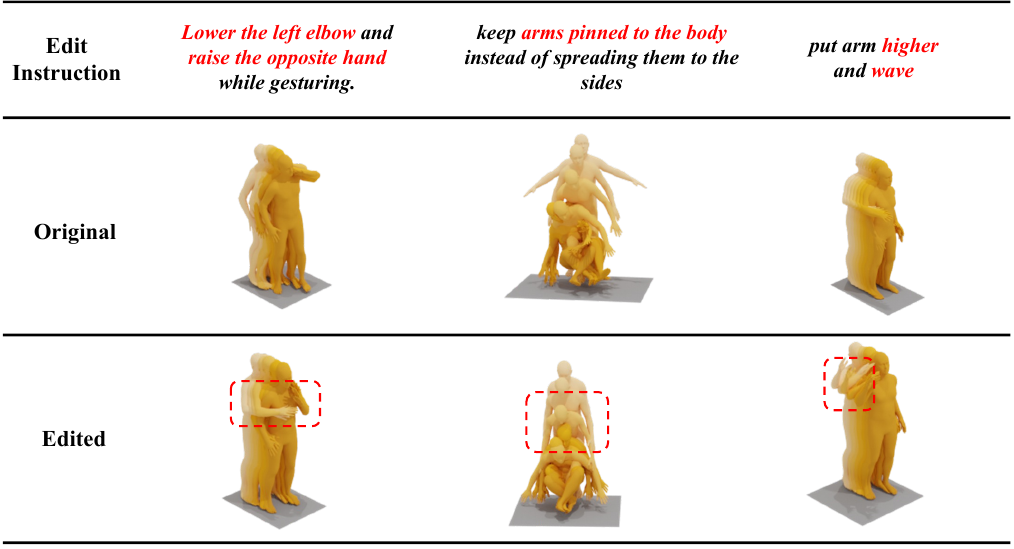}
    \vspace{-2mm}
    \caption{Qualitative comparison of text-conditioned motion editing. Each column shows the editing instruction, the original motion, and the edited result from UniMotion. Red text highlights the target edit attributes in the instruction, and red dashed boxes mark the body regions primarily affected by the edit. UniMotion executes the requested change more accurately while better preserving the original motion content outside the edited regions.}
    \label{fig:suppl_edit}
    \vspace{-2mm}
\end{figure}

Figure~\ref{fig:suppl_edit} presents qualitative results on MotionFix-style editing instructions. UniMotion performs targeted motion modification while retaining the unedited content of the source sequence. The edits are more localized and semantically precise, particularly for instructions involving limb-specific changes, speed modulation, and posture refinement.

\subsection{Vision-to-Motion}

\begin{figure}[htbp]
    \centering
    \includegraphics[width=\linewidth]{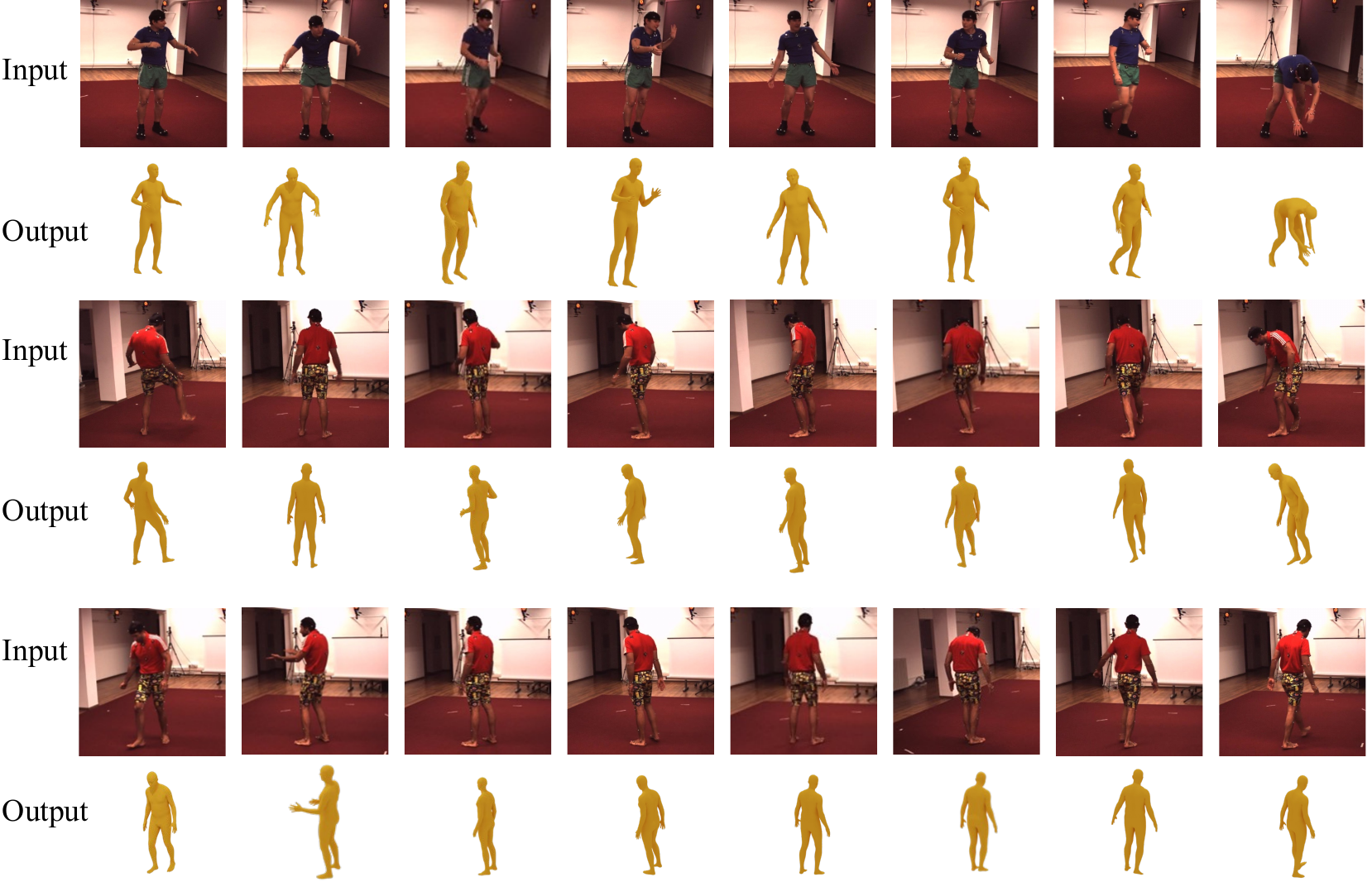}
    \vspace{-2mm}
    \caption{Qualitative Vision-to-Motion results. For each example, we show a sequence of sampled visual inputs together with the corresponding motion outputs recovered by UniMotion. The predicted body structure remains consistent across observations, with accurate limb orientation, stable torso alignment, and clear whole-body geometry.}
    \label{fig:suppl_rgb2m_qual}
    \vspace{-2mm}
\end{figure}

Figure~\ref{fig:suppl_rgb2m_qual} visualizes representative Vision-to-Motion results on Human3.6M. UniMotion recovers body structure with more accurate limb orientation and torso alignment, while maintaining strong consistency across sampled visual observations. The qualitative gains are most apparent in asymmetric poses and cases with large arm articulation.

\subsection{Vision-to-Text: Image and Video Inputs}
\label{sec:suppl_rgb_grounded_qual}

\begin{figure}[t]
    \centering
    \includegraphics[width=\linewidth]{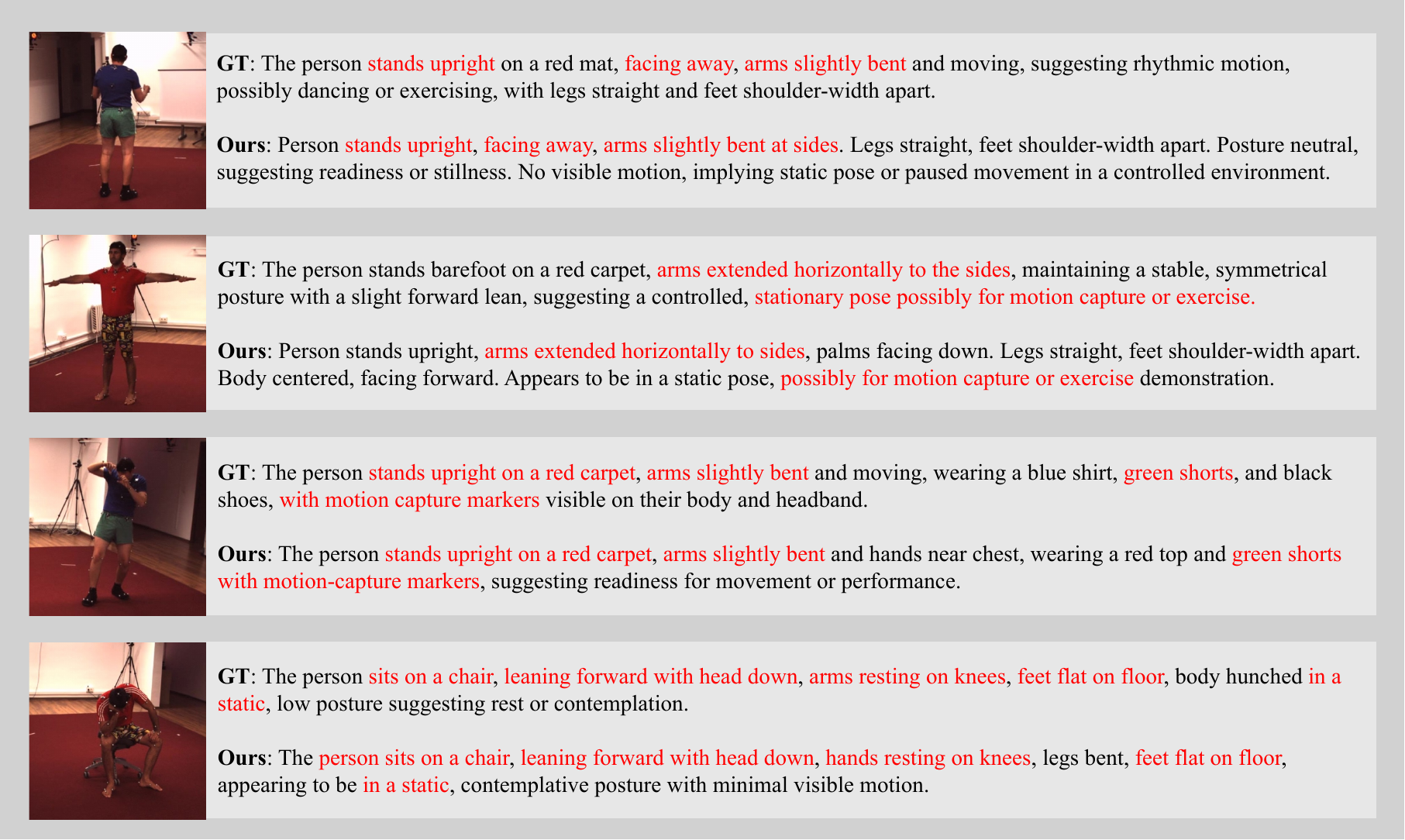}
    \vspace{-2mm}
    \caption{Qualitative Vision-to-Text results on H3.6M. Each example shows the visual input together with the ground-truth description and the caption generated by UniMotion. Red phrases highlight the key pose-aware semantics shared by the reference and the prediction. UniMotion converts visual input into concise language that accurately describes body configuration, limb orientation, and local articulation from static visual evidence.}
    \label{fig:suppl_rgb2t_qual}
    \vspace{-2mm}
\end{figure}

\begin{figure}[htbp]
    \centering
    \includegraphics[width=\linewidth]{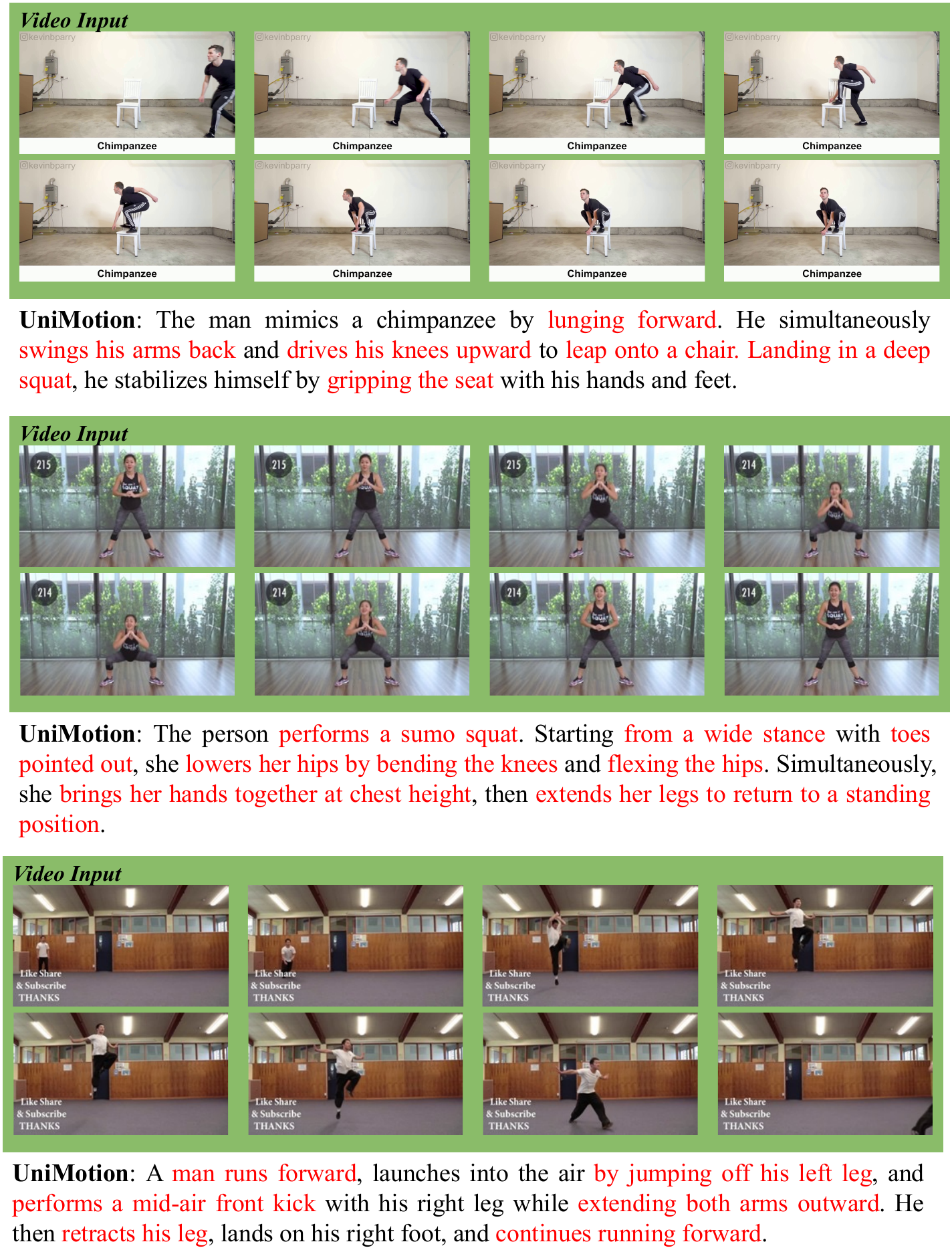}
    \vspace{-2mm}
    \caption{Qualitative Vision-to-Text results on MoVid (video inputs). Each example shows multiple sampled video frames followed by the description generated by UniMotion, with red phrases highlighting the key temporal action components. Given sparse frame observations, UniMotion produces temporally coherent descriptions of the full action, capturing action category, posture evolution, and phase transitions.}
    \label{fig:suppl_movid_qual}
    \vspace{-2mm}
\end{figure}

Vision-to-Text spans two temporal regimes under the same vision-conditioned language interface: single-frame pose description on H3.6M, and multi-frame dynamic action understanding on MoVid. Figures~\ref{fig:suppl_rgb2t_qual} and~\ref{fig:suppl_movid_qual} present these two settings separately: the first focuses on H3.6M Vision-to-Text from a single image, while the second shows how the same interface scales to video-level Vision-to-Text on MoVid.

In Figure~\ref{fig:suppl_rgb2t_qual}, UniMotion consistently captures static body configuration cues that are visually prominent in the input frame, including facing direction, arm extension, seated vs.\ standing posture, and coarse limb placement. The examples show that the model preserves the core pose semantics even when the generated wording is not identical to the reference, which is particularly important for human-centric visual understanding where multiple valid descriptions may exist for the same pose.

Figure~\ref{fig:suppl_movid_qual} shows that the same interface extends beyond static pose description to dynamic action understanding from sparsely sampled video frames. In these examples, UniMotion not only identifies the action category (\eg, animal-mimicking motion, squat, and running-jump sequences), but also reconstructs temporally ordered sub-actions such as lunging, lowering the hips, airborne kicking, and landing. This suggests that the model is not merely recognizing isolated poses, but inferring the motion phase transition across frames.

\subsection{Motion Reconstruction Quality}

We systematically compare the reconstruction quality of CMA-VAE, VQ-VAE (MotionGPT~\cite{motiongpt}), and MLD-VAE (MLD~\cite{mld}) through four complementary analyses spanning the statistical, spatial, frequency, and temporal domains. All visualizations are computed on the \textbf{common sample set}---the intersection of successfully reconstructed samples across all three models on the HumanML3D test split---ensuring strictly fair comparison. Joint positions follow the standard HumanML3D 22-joint convention; all errors are reported in millimeters (mm). As reported in Table~6 of the main paper, CMA-VAE achieves the best reconstruction across all metrics (APE=3.53\,cm, AVE=0.428\,cm/s, FID=0.0282), substantially outperforming both VQ-VAE (APE=17.15, AVE=0.813, FID=0.0674) and MLD-VAE (APE=9.28, AVE=0.981, FID=0.1283). The following visualizations provide deeper insight into the nature of these differences.

\subsubsection{Dataset-Level Error Distribution}

\begin{figure}[htbp]
    \centering
    \includegraphics[width=\linewidth]{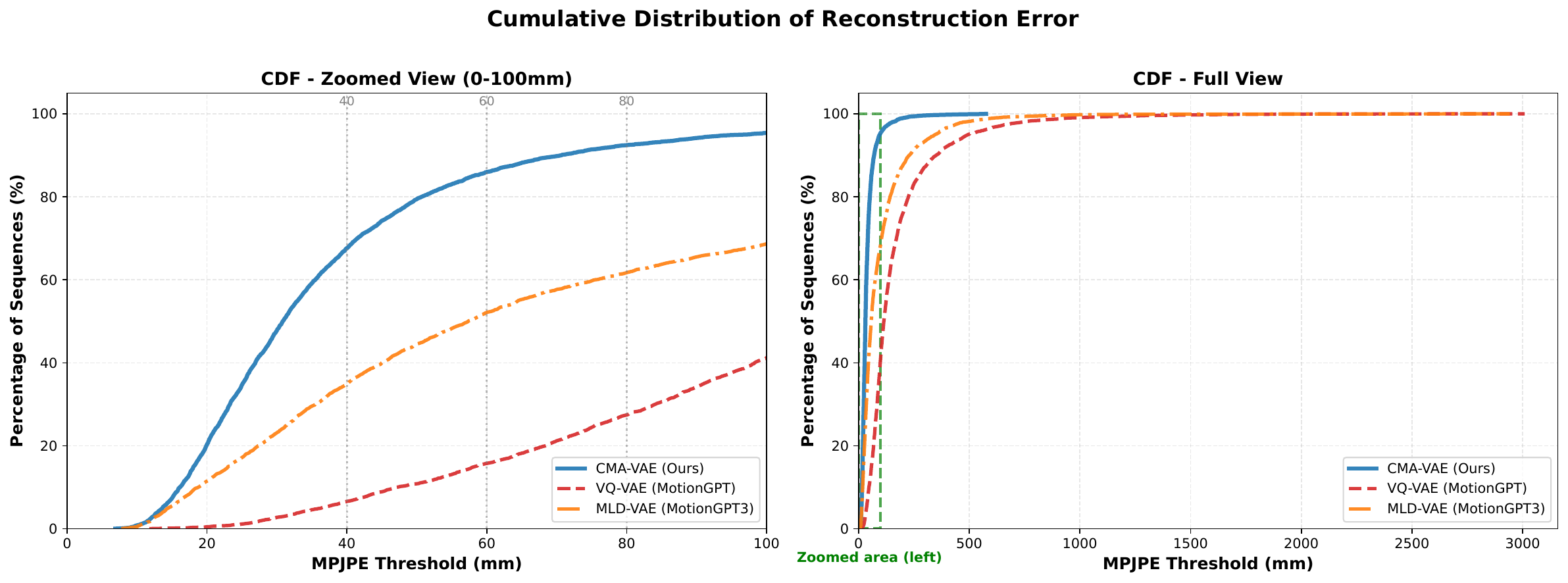}
    \vspace{-3mm}
    \caption{Cumulative distribution of per-sequence reconstruction MPJPE for CMA-VAE, VQ-VAE, and MLD-VAE. \textbf{Left}: zoomed view (0--100\,mm) with reference thresholds at 40, 60, and 80\,mm. \textbf{Right}: full-range view (0--3000\,mm), where the green dashed box indicates the region enlarged in the left panel. CMA-VAE's CDF is uniformly left-shifted, indicating lower reconstruction error across the entire test set. VQ-VAE exhibits a long heavy tail ($>$2000\,mm), reflecting catastrophic quantization failures on a subset of sequences.}
    \label{fig:suppl_vae_cdf}
    \vspace{-2mm}
\end{figure}

Figure~\ref{fig:suppl_vae_cdf} shows the cumulative distribution function (CDF) of per-sequence MPJPE. The left panel zooms into the 0--100\,mm range where the majority of CMA-VAE sequences concentrate, while the right panel reveals the full distributional picture up to 3000\,mm. CMA-VAE's curve is uniformly left-shifted: at a 40\,mm threshold, approximately 70\% of CMA-VAE sequences fall below it, compared to $\sim$30\% for VQ-VAE and $\sim$35\% for MLD-VAE. In the full-range view, VQ-VAE exhibits a prominent heavy tail extending beyond 2000\,mm, whereas CMA-VAE converges almost entirely within 600\,mm. This confirms that CMA-VAE's advantage is a \textit{dataset-wide} distributional shift rather than gains on a narrow subset of easy samples.

\subsubsection{Joint-wise Spatial Error Analysis}

\begin{figure}[htbp]
    \centering
    \includegraphics[width=0.8\linewidth]{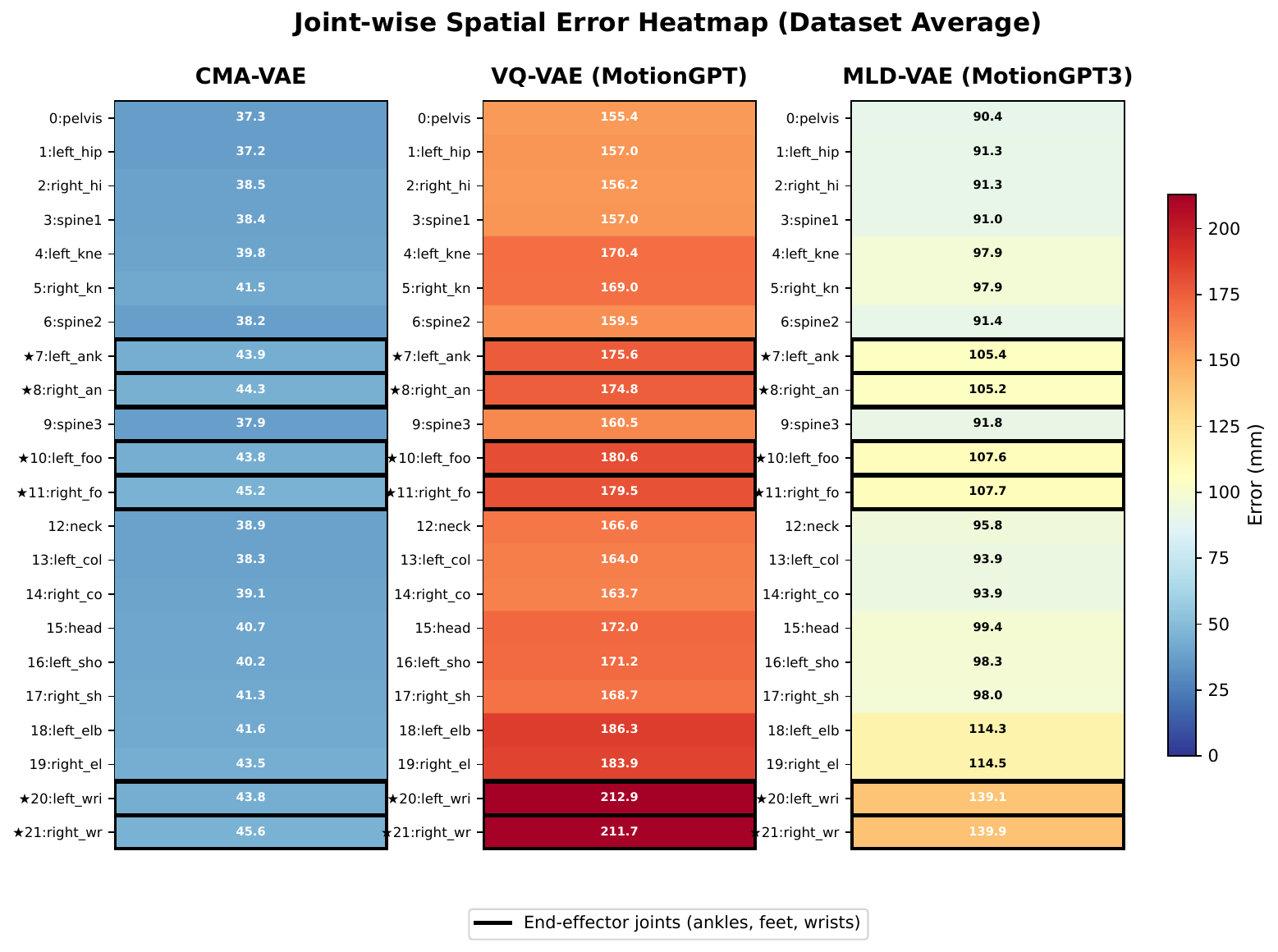}
    \vspace{-3mm}
    \caption{Joint-wise spatial error heatmap (dataset average, mm). Each column shows one model, and rows correspond to the 22 HumanML3D joints, with the numerical error annotated in each cell. Color encodes error magnitude (blue = low, red = high), and end-effector joints ($\star$) are highlighted with black borders. CMA-VAE maintains uniformly low error across all joints, while VQ-VAE and MLD-VAE exhibit disproportionately elevated end-effector errors.}
    \label{fig:suppl_vae_heatmap}
    \vspace{-2mm}
\end{figure}

Figure~\ref{fig:suppl_vae_heatmap} provides a per-joint breakdown of the dataset-averaged reconstruction error. Each column corresponds to one model, with all 22 joints listed vertically and the numerical error (mm) annotated. End-effector joints (ankles, feet, wrists, marked with $\star$) are highlighted with black borders, as they accumulate the largest kinematic-chain errors.

Three key observations emerge: (1)~CMA-VAE achieves uniformly low error across all joints (37--46\,mm), with minimal variation between trunk and end-effectors---evidence that the continuous latent space preserves both global pose structure and distal articulation fidelity. (2)~VQ-VAE exhibits systematically elevated error (155--213\,mm), with wrists suffering the most (left/right wrist: 212.9/211.7\,mm), confirming that discrete codebook quantization disproportionately degrades high-amplitude end-effector dynamics. (3)~MLD-VAE falls between the two (90--140\,mm), but its wrist errors (139.1/139.9\,mm) are notably higher than its trunk errors ($\sim$91\,mm), indicating that a plain continuous VAE without cross-modal anchoring still struggles with end-effector fidelity.

\subsubsection{Residual Frequency Spectrum Analysis}

To complement the spatial analyses above, we examine reconstruction fidelity in the \textit{frequency domain}. For each sequence, we compute the residual signal $\mathbf{r}_{j,t} = \mathbf{p}_{j,t}^{\text{pred}} - \mathbf{p}_{j,t}^{\text{gt}}$ across all joints and coordinate axes, apply the FFT, and average the resulting power spectra over up to 1000 randomly sampled sequences. The frequency axis ranges from 0 to the Nyquist limit of 10\,Hz (HumanML3D operates at 20\,fps), partitioned into three physically meaningful bands: \textit{low} (0--2\,Hz), capturing gross locomotion; \textit{mid} (2--6\,Hz), corresponding to limb swing and gait cycles; and \textit{high} (6--10\,Hz), encoding rapid articulation and contact dynamics.

\begin{figure}[htbp]
    \centering
    \includegraphics[width=0.8\linewidth]{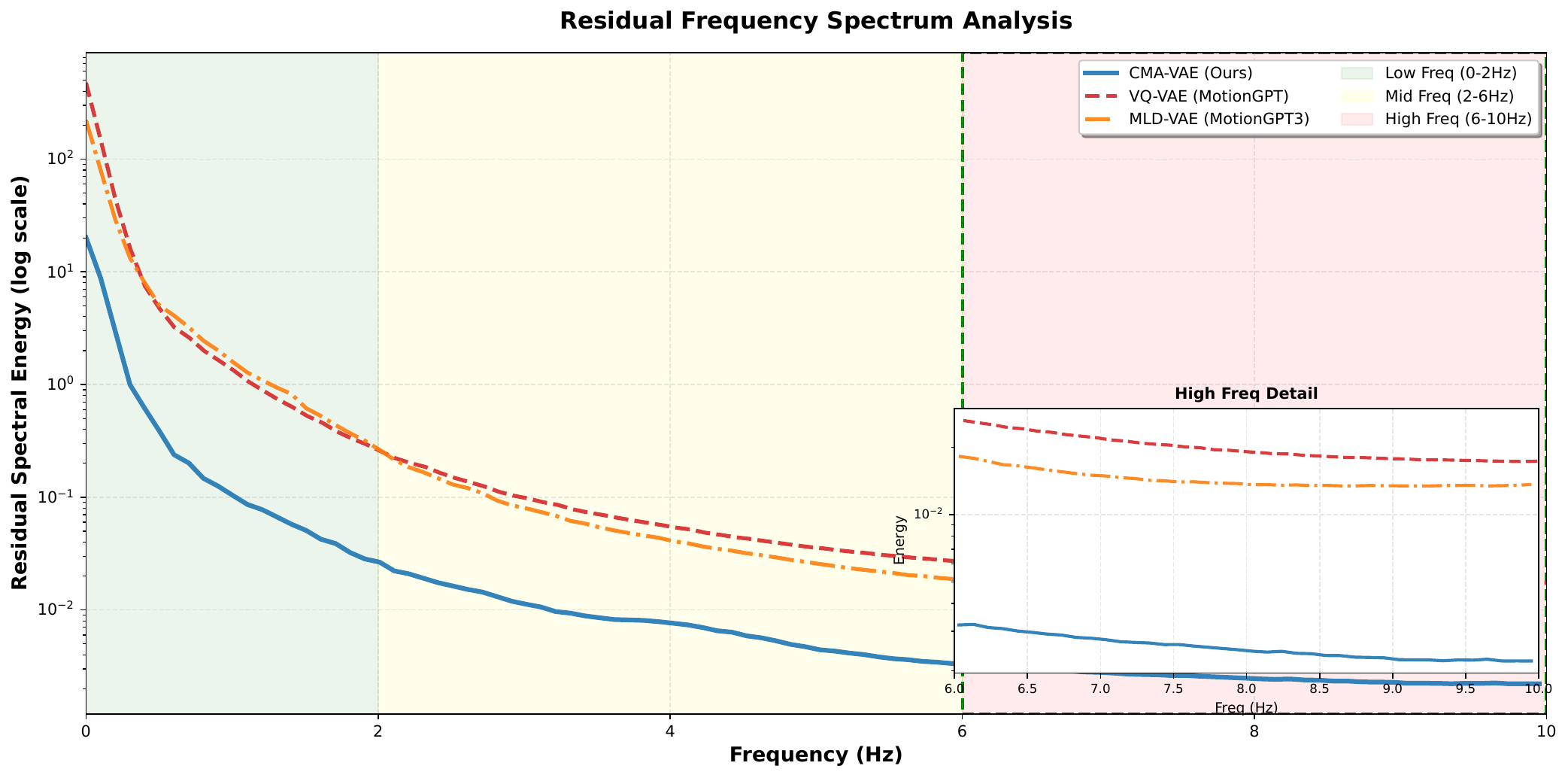}
    \vspace{-3mm}
    \caption{Residual frequency spectrum analysis. The main plot shows the mean residual spectral energy (log scale) versus frequency (Hz), with background shading denoting low (0--2\,Hz), mid (2--6\,Hz), and high (6--10\,Hz) bands. The inset provides a magnified view of the high-frequency range (6--10\,Hz). CMA-VAE (blue) maintains the lowest residual energy across all bands, while VQ-VAE (red) exhibits elevated high-frequency residuals from discrete codebook quantization artifacts.}
    \label{fig:suppl_vae_spectrum}
    \vspace{-2mm}
\end{figure}

As shown in Figure~\ref{fig:suppl_vae_spectrum}, CMA-VAE achieves the lowest residual spectral energy across \textit{all} frequency bands, with approximately one order of magnitude separation from VQ-VAE throughout the mid- and high-frequency ranges. The high-frequency inset (6--10\,Hz) is particularly revealing: VQ-VAE's residual energy plateaus at $\sim$2$\times$10$^{-2}$, indicating that discrete codebook switching injects spurious high-frequency artifacts into the reconstructed motion---precisely the ``temporal jitter'' described qualitatively in the main paper. MLD-VAE shows moderately lower high-frequency residuals than VQ-VAE but remains substantially above CMA-VAE, consistent with its intermediate reconstruction quality. CMA-VAE's uniformly low spectral residuals confirm that DPA-calibrated continuous representations faithfully preserve motion dynamics at all temporal scales.

\subsubsection{Temporal Velocity Fidelity and Jitter Analysis}

\begin{figure}[t]
    \centering
    \includegraphics[width=\linewidth]{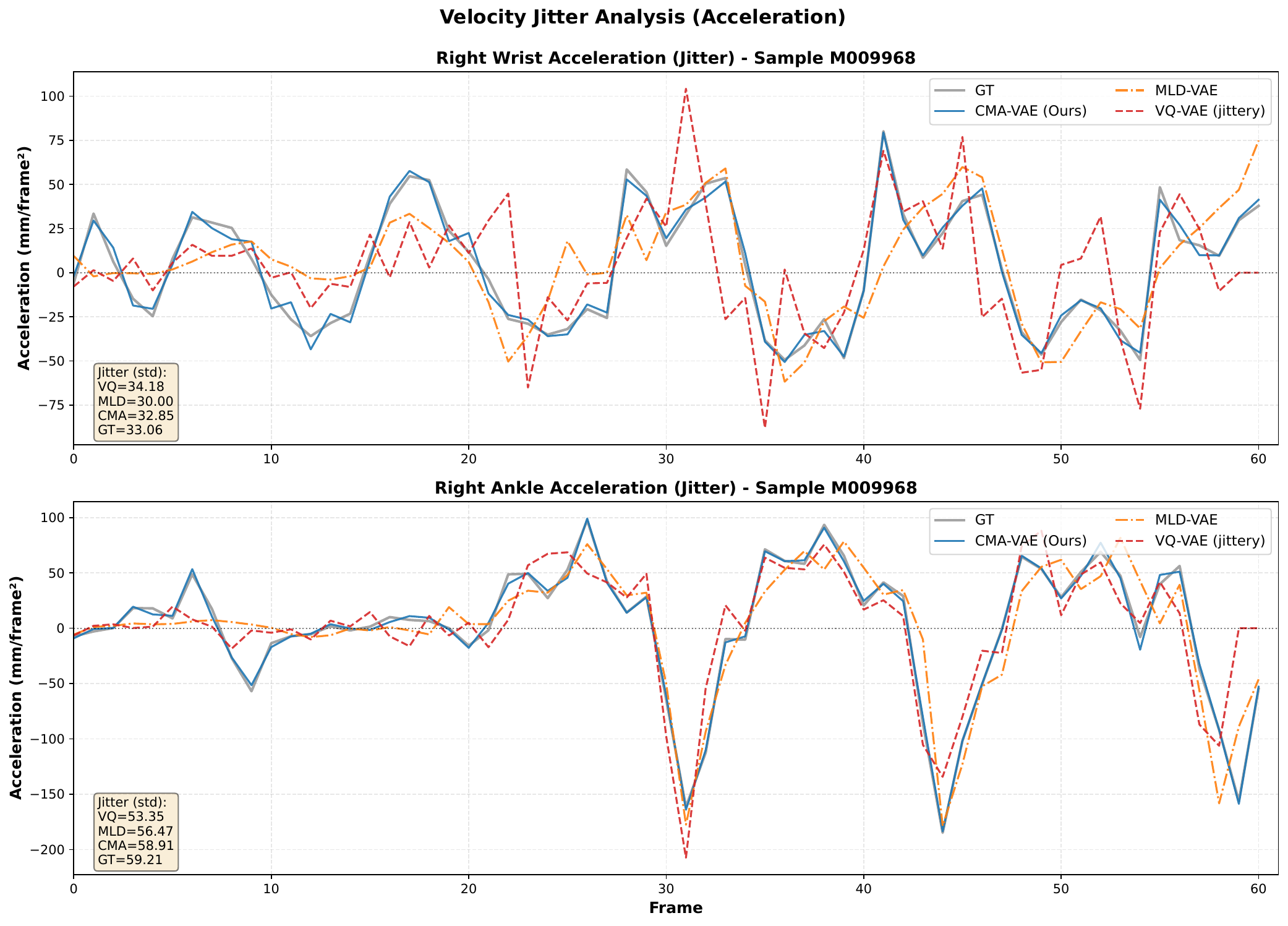}
    \vspace{-3mm}
    \caption{Acceleration-based jitter analysis on a high-dynamic sample (M009968). \textbf{Top}: right wrist acceleration (mm/frame$^2$). \textbf{Bottom}: right ankle acceleration. Each panel overlays GT (gray), CMA-VAE (blue), MLD-VAE (orange), and VQ-VAE (red), and the inset box reports the acceleration standard deviation used as the jitter metric. CMA-VAE most closely matches the ground-truth dynamics, while VQ-VAE introduces high-frequency jitter and MLD-VAE over-smooths peak accelerations.}
    \label{fig:suppl_vae_jitter}
    \vspace{-2mm}
\end{figure}

Finally, we quantify temporal smoothness by analyzing the acceleration signal (second-order finite difference of joint position), whose standard deviation serves as a direct measure of motion jitter. Figure~\ref{fig:suppl_vae_jitter} plots the per-frame acceleration for the right wrist and right ankle on a high-dynamic sample (M009968, selected from the top-5 fastest sequences by 95th-percentile velocity).

CMA-VAE's acceleration profile closely tracks the ground truth: on the right wrist, jitter std = 32.85 vs.\ GT = 33.06 (0.6\% relative deviation); on the right ankle, jitter std = 58.91 vs.\ GT = 59.21 (0.5\% deviation). In contrast, VQ-VAE produces noticeably jittery acceleration with higher peak magnitudes, reflecting abrupt code-switching during rapid transitions. MLD-VAE exhibits over-smoothed acceleration with systematically reduced peak amplitudes, consistent with its higher AVE (0.981\,cm/s) caused by temporal detail loss. These results provide direct time-domain evidence that CMA-VAE preserves both the \textit{magnitude} and \textit{temporal structure} of motion dynamics with near-ground-truth fidelity.

\section{Further Analysis}
\label{sec:suppl_analysis}

\subsection{DPA Stability and Reverse KL Analysis}
\label{sec:suppl_dpa_analysis}

\noindent\textbf{Reverse KL direction: mode-seeking distillation.}
The DPA alignment loss adopts $D_\mathrm{KL}(q_\phi(z \mid \mathbf{m}) \| q_\psi(z \mid \mathbf{m}, \mathbf{v}))$ with $q_\psi$ \textbf{detached} (stop-gradient) as the alignment target. In the knowledge-distillation convention where $q_\psi$ (vision-fused) is the teacher and $q_\phi$ (motion-only) is the student, this is the \emph{reverse} KL (student$\to$teacher). Minimizing $D_\mathrm{KL}(q_\phi \| q_\psi)$ w.r.t.\ $q_\phi$ is \emph{mode-seeking}: $q_\phi$ is penalized heavily for placing probability mass where $q_\psi$ assigns low density (the $\log q_\psi$ term in the KL diverges toward $-\infty$), driving $q_\phi$ to concentrate on the most prominent semantic modes of the vision-fused posterior. This yields compact, high-confidence motion representations that capture the core visual-semantic information while naturally filtering out view-specific or noise-related visual modes irrelevant to motion semantics.

In contrast, the forward KL $D_\mathrm{KL}(q_\psi \| q_\phi)$ would be \emph{mode-covering}: optimizing $q_\phi$ under this objective would force it to assign probability mass wherever $q_\psi$ has support---including noisy or view-dependent visual modes---resulting in an over-dispersed motion posterior. For our diagonal-Gaussian posteriors, this manifests as overestimated variance in $q_\phi$, diluting representational precision. The reverse KL instead produces tighter variance, which is preferable since the motion encoder at inference should yield precise, confident encodings without image input.

\noindent\textbf{Training dynamics and stability.}
The DPA loss $\mathcal{L}_\mathrm{align}$ uses a linear warm-up schedule over the first 10k training steps. This prevents the alignment constraint from dominating during early training when the VAE's basic reconstruction ability is still unstable. As training progresses, $\mathcal{L}_\mathrm{align}$ gradually increases while $\mathcal{L}_\mathrm{recon}$ and $\mathcal{L}_\mathrm{KL}$ continue to decrease---the three losses converge cooperatively rather than competing. The posterior variances $\mathbb{E}[\sigma^2_\phi]$ and $\mathbb{E}[\sigma^2_\psi]$ remain in a healthy range (not collapsing toward zero) throughout training, confirming no posterior collapse occurs.

\noindent\textbf{Generalization to unpaired datasets.}
HumanML3D data does not participate in $\mathcal{L}_\mathrm{align}$ (no paired images). However, because the Motion Encoder's front-end parameters are shared with the Vision-Fused Encoder, DPA's gradients from H36M data indirectly improve representations for HumanML3D as well. This explains why removing DPA degrades T2M performance (R@3: 0.841 $\to$ 0.818), even though T2M evaluation uses only HumanML3D data. The shared-parameter design enables implicit knowledge transfer across datasets with different modality coverage.

\subsection{LRA Non-Triviality Analysis}
\label{sec:suppl_lra_analysis}

A natural concern is whether LRA's M2M self-reconstruction degenerates into a trivial identity mapping. We provide three layers of evidence that this is not the case.

\noindent\textbf{(1) Architectural impossibility.}
Three non-trivial transformations separate the conditioning input from the reconstruction target:
\begin{itemize}
    \item \textbf{Dimension remapping}: $z \in \mathbb{R}^{T_z \times 256}$ is projected and compressed by the dual-path embedder into LLM tokens of dimension $d_h{=}1536$, involving non-linear MLP projections and 4-layer Transformer encoding in the semantic branch.
    \item \textbf{LLM processing}: The LLM backbone operates on the fused tokens with causal attention and LoRA adaptation---a highly non-linear transformation.
    \item \textbf{Probabilistic flow matching}: The flow head predicts velocity fields starting from pure Gaussian noise $z_0 \sim \mathcal{N}(0,I)$, conditioned on LLM hidden states via AdaLN. The prediction target depends on both the noise realization and the timestep $t$---the model must ``understand'' motion structure to produce correct velocity directions, rather than copying input features.
\end{itemize}

\noindent\textbf{(2) Information bottleneck during training.}
Beyond architectural constraints, we apply explicit information bottleneck regularization to the conditioning input during M2M training:
\begin{itemize}
    \item \textbf{Temporal subsampling}: Randomly retain only 20--50\% of temporal frames from the motion latent conditioning, requiring the model to interpolate missing temporal information.
    \item \textbf{Feature dropout}: Randomly zero out 15\% of feature dimensions in the conditioning tokens.
    \item \textbf{Gaussian perturbation}: Add noise with $\sigma{=}0.02$ to the conditioning latents.
\end{itemize}
The model receives a \emph{degraded} version of $z$ as conditioning but must reconstruct the \emph{complete, clean} target $z$---identity mapping is strictly impossible since input $\neq$ target. This design is analogous to masked image modeling and denoising autoencoders, which are well-established as non-trivial self-supervised objectives.

\noindent\textbf{(3) Cross-task downstream transfer.}
The strongest evidence against trivial memorization is cross-task generalization: the M2M-calibrated pathway achieves T2M R@3=0.841 (vs.\ 0.801 without LRA) and simultaneously improves Vision$\to$M MPJPE (84.3$\to$75.0), even though M2M training involves \emph{no text or image inputs}. An identity mapping of motion latents cannot explain gains on text-driven generation or image-driven estimation---the pathway must have learned transferable structural representations of motion dynamics.

\noindent\textbf{Shuffled-condition control experiment.}
To directly verify that the model learns condition-target correspondence rather than dataset-level statistics, we evaluate the trained LRA model with mismatched conditions: using motion A's embedding to guide reconstruction of motion B's target. Under matched conditions, reconstruction FID is 0.008; under shuffled conditions, FID degrades to 2.34 ($\sim$300$\times$ worse), confirming that the model has learned a structured condition$\to$target mapping rather than a generic motion prior.

\subsection{Out-of-Distribution Generalization}

To evaluate generalization beyond the training distribution, we test UniMotion's Vision-to-Motion capability on 3DPW~\cite{3dpw} without any fine-tuning. Despite training exclusively on H36M (indoor, controlled setting), UniMotion achieves competitive performance on 3DPW's in-the-wild scenarios:

\begin{table}[htbp]
\centering
\small
\renewcommand\arraystretch{1.1}
\caption{Zero-shot Vision-to-Motion on 3DPW (no fine-tuning).}
\vspace{-1mm}
\label{tab:ood_3dpw}
\begin{tabular}{@{}lcc@{}}
\toprule
\textbf{Method} & \textbf{MPJPE$\downarrow$} & \textbf{PA-MPJPE$\downarrow$} \\
\midrule
UniPose~\cite{unipose} (zero-shot) & 99.4 & 65.8 \\
\rowcolor{mygray}
\textbf{UniMotion (zero-shot)} & \textbf{93.6} & \textbf{58.3} \\
\bottomrule
\end{tabular}
\vspace{-2mm}
\end{table}

This demonstrates that DPA's visual-motion alignment generalizes beyond the training domain. The relative improvement over UniPose is consistent with in-domain results, suggesting that CMA-VAE's continuous representations capture transferable body structure priors rather than overfitting to H36M-specific visual patterns.

\section{Architecture and Implementation Details}
\label{sec:suppl_arch}

\subsection{Unified 269-Dimensional Motion Representation}

A core issue in prior work is representational inconsistency across tasks. Text-driven motion generation typically uses HumanML3D-style 263-dimensional representations, while image-driven body recovery operates directly in the raw SMPL parameter space. As a result, motion generation and body recovery conventionally use two separate interfaces, making end-to-end unified training within a single framework difficult.

UniMotion adopts a single \textbf{269-dimensional} representation to bridge this gap. This representation maintains backward compatibility with the conventional 263-dim representation and additionally introduces 6-dim global orientation information to support body-structure supervision and cross-modal alignment.

For each frame, we define a 269-dimensional vector:
\begin{enumerate}
    \item Root motion increment: 3-dim (1-dim rotation increment + 2-dim planar translation increment).
    \item Root height: 1-dim.
    \item Relative joint positions: 63-dim ($21$ joints $\times$ 3).
    \item Local joint rotations (continuous 6D form): 126-dim ($21$ joints $\times$ 6).
    \item Local joint velocities: 66-dim ($22$ joints $\times$ 3).
    \item Foot contact states: 4-dim.
    \item \textbf{Global orientation (continuous 6D form): 6-dim.}
\end{enumerate}
The first six components sum to 263-dim, consistent with mainstream motion generation representations. The newly added 6-dim component maintains global orientation consistency in visual-motion tasks. The existing HumanML3D evaluation protocol can be applied directly to the first 263 dimensions.

\noindent\textbf{Construction per dataset.}
\textbf{HumanML3D}: We strictly preserve the official 263-dim features unchanged, appending only the 6-dim global orientation computed under the same kinematic conventions.
\textbf{MotionFix}: We first align the coordinate system and joint conventions to a unified semantic, then construct sequence-level 269-dim features in both time-difference mode (compatible with HumanML3D) and frame-level SMPL-semantic consistent mode.
\textbf{Human3.6M}: Data is converted from SMPL annotations to the same 22-joint convention, then encoded into 269-dim features with unified coordinate normalization.

\subsection{CMA-VAE Architecture Details}
\label{sec:suppl_cmavae}

\noindent\textbf{Motion-guided joint sampling.}
The 2D joint positions $\mathbf{j}_{2d}(\mathbf{m})$ used for grid sampling in the Vision-Fused Encoder are derived from motion in one of three ways, evaluated in priority order:
\begin{enumerate}
    \item \textbf{Precomputed skeleton projections} (primary path, used for H36M): 2D joint coordinates are pre-projected from ground-truth 3D skeleton annotations under full camera calibration and stored offline. This provides the most geometrically precise sampling grid.
    \item \textbf{Dataset-provided image-space annotations} (\texttt{camera\_params[`joint3d\_image']}): pixel-space joint coordinates from dataset annotations, used when precomputed files are unavailable.
    \item \textbf{Runtime forward-kinematics recovery} (fallback): 3D joint positions are recovered from the 269-dim motion representation via \texttt{recover\_from\_ric} and projected using weak-perspective camera parameters.
\end{enumerate}
All three paths share the key property that the grid-sampling positions are \emph{structurally grounded in the motion skeleton}: the same joint coordinates that define the motion representation also determine which image regions are attended to. This motion-guided coupling is what gives the Vision-Fused Encoder genuine cross-modal sensitivity---visual features are extracted specifically at body-relevant locations rather than global image regions.

\noindent\textbf{Reference-frame visual guidance.}
Since H36M training samples consist of individual frames paired with motion sequences, the Vision-Fused Encoder extracts visual features from the reference frame and broadcasts them across the full temporal sequence.
This design is justified by the DPA objective: DPA aligns the \emph{distributional statistics} of the two encoder posteriors, which requires injecting body-identity and scene-context semantics rather than per-frame visual detail. A single reference frame is sufficient to supply this semantic supervision, while per-frame extraction would multiply HRNet's cost by sequence length ($\sim$60--200 frames) with no commensurate benefit to the distributional alignment objective.

\subsection{Unified Architecture: Detailed Formulations}
\label{sec:suppl_arch_detail}

This section provides the full architectural details for the components introduced in Sec.~3.3 of the main paper.

\noindent\textbf{Dual-Path Embedder.}
Given the CMA-VAE latent $z \in \mathbb{R}^{T_z \times d}$, the embedder contains two parallel branches. The \textbf{Semantic branch} maps $z$ to semantic feature dimension $d_s{=}512$ via MLP; after adding learnable positional encodings, $N_s{=}4$ Transformer Encoder layers extract high-level semantic features:
\begin{equation}
    \mathbf{e}_\mathrm{und} = \mathrm{TransformerEnc}\bigl(\mathrm{MLP}(z) + \mathrm{PosEmbed}\bigr) \in \mathbb{R}^{T_z \times d_s}.
\end{equation}
This branch mirrors the SigLIP Encoder on the vision side and captures global motion semantics. The \textbf{Generation branch} directly maps $z$ to the LLM hidden dimension $d_h$ via MLP, with independent learnable positional encodings preserving low-level motion details:
\begin{equation}
    \mathbf{e}_\mathrm{gen} = \mathrm{MLP}(z) + \mathrm{PosEmbed}_\mathrm{gen} \in \mathbb{R}^{T_z \times d_h}.
\end{equation}
This branch mirrors the PatchEmbed on the vision side. The two branch outputs are concatenated along the channel dimension and projected to the unified LLM hidden dimension via RMSNorm + MLP:
\begin{equation}
    \mathbf{e}_\mathrm{fused} = \mathrm{FusionProj}\bigl([\mathbf{e}_\mathrm{und} \,\|\, \mathbf{e}_\mathrm{gen}]\bigr) \in \mathbb{R}^{T_z \times d_h}.
\end{equation}
All tasks (understanding and generation) uniformly use fused embeddings, consistent with the image modality.

\noindent\textbf{Vision Pathway and Pose-Aware Vision Backbone.}
For visual inputs, UniMotion reuses Show-o2's native image processing pipeline. Input images are first encoded by the pre-trained WAN2.1 VAE~\cite{wan2025wan} into continuous latents $\mathbf{l}_\mathrm{img} \in \mathbb{R}^{C \times H' \times W'}$ ($C{=}16$, spatial compression factor $8{\times}$), then processed through the symmetric dual-path design: the Semantic Branch uses PatchEmbed followed by SigLIP Vision Encoder; the Generation Branch uses an independent PatchEmbed to preserve spatial details. The two branches are fused via FusionProj before entering the LLM. For 432$\times$432 input resolution, the WAN2.1 VAE produces 54$\times$54 latents, and with patch\_size=2, this yields 27$\times$27=729 image tokens---exactly matching the SigLIP position encoding.

To provide fine-grained human body structure understanding for human-centric tasks, the RGB pathway is additionally equipped with a \textbf{pose-aware vision backbone}. This backbone employs a ViT-H backbone (1280-dim, 32 layers, 16 heads) initialized from a pretrained human body encoder~\cite{tokenhmr} and kept \textbf{frozen} throughout training. Given an RGB image, it extracts body-structure-aware spatial features $\mathbf{f}_\mathrm{pose} \in \mathbb{R}^{d_p \times h_p \times w_p}$ ($d_p{=}1280$), which are bilinearly interpolated to match the image token spatial resolution ($h'{=}w'{=}27$) and flattened into $N{=}729$ tokens. These pose-aware features are then concatenated with the fused SigLIP+PatchEmbed image embeddings and re-projected via a lightweight fusion layer:
\begin{equation}
    \mathbf{e}_\mathrm{rgb} = \mathrm{PoseFusionProj}\bigl([\mathbf{e}_\mathrm{img\_fused} \,\|\, \mathbf{f}_\mathrm{pose}]\bigr) \in \mathbb{R}^{N \times d_h},
\end{equation}
where PoseFusionProj is RMSNorm($d_h + d_p$) $\to$ Linear $\to$ GELU $\to$ Linear.

This design maintains architectural symmetry with the Motion side: the Motion dual-path embedder provides both semantic (global) and generation (detail-preserving) features, and the visual pathway similarly combines SigLIP's general visual semantics with the pose-aware vision backbone's human-centric geometric features. Quantitatively, this combination yields the best Vision$\to$M performance reported in Table~7, while Table~9 shows that the aligned motion pathway already provides a strong base before the final visual refinements.

\noindent\textbf{Hybrid Attention.}
We design a hybrid attention mechanism that maintains \textbf{global causal} constraints at the sequence level while applying \textbf{intra-motion full attention} within motion token spans. Formally, given a mixed sequence of length $L$ comprising text tokens $\mathcal{T}$, image tokens $\mathcal{I}$, and the $k$-th motion span $\mathcal{M}_k$, the attention mask $\mathbf{M} \in \{0, -\infty\}^{L \times L}$ is defined as:
\begin{equation}
M_{ij} = \begin{cases}
0 & \text{if } i \in \mathcal{T},\; j \leq i \\
0 & \text{if } i \in \mathcal{M}_k,\; j \in \mathcal{M}_k \\
0 & \text{if } i \in \mathcal{M}_k,\; j \notin \mathcal{M}_k,\; \mathrm{pos}(j) < \mathrm{start}(\mathcal{M}_k) \\
0 & \text{if } i \in \mathcal{I},\; j \in \mathcal{I} \\
0 & \text{if } i \in \mathcal{I},\; j \notin \mathcal{I},\; \mathrm{pos}(j) < \mathrm{start}(\mathcal{I}) \\
-\infty & \text{otherwise}
\end{cases}
\end{equation}

\noindent\textbf{No information leakage guarantee.} Motion and image tokens share the same attention pattern: each can \emph{only} attend to (1) all tokens within its own span (bidirectional full attention), and (2) all tokens \emph{preceding} that span (unidirectional). They \emph{cannot} attend to any token beyond the span boundary. Text tokens follow standard causal attention ($j \leq i$). Consequently, when generating the $i$-th text token, it sees only positions $\leq i$ and all completed motion/image spans before it. No future information leaks into the generation process. This design reconciles the flow matching objective---which requires simultaneous velocity field prediction across the entire motion or image latent sequence---with text's autoregressive constraints.

\noindent\textbf{Modality-Routed LoRA.}
For each $Q$, $K$, $V$, $O$ projection matrix in every attention layer, we attach two Low-Rank Adaptation branches: \textbf{LoRA-A} for Text/RGB tokens and \textbf{LoRA-B} for Motion tokens, each with rank 32. During the forward pass, a deterministic modality mask routes each token to its corresponding LoRA branch based on modality identity (always known from sequence construction), eliminating the gating uncertainty of learned MoE routing~\cite{hmvlm}. This introduces only $\sim$2\% additional parameters while enabling modality-specific parameter adaptation. Unlike HMVLM's MoE-LoRA which requires load-balancing losses and introduces routing noise during training, our deterministic design is more appropriate for the tri-modal setting where modality identity is unambiguous.

\subsection{Flow Matching Generation}
\label{sec:suppl_flow}

UniMotion adopts the flow matching framework~\cite{flowmatching} for Motion and RGB generation. Given a data point $x_1$ (target Motion Latent or Image Latent), noise $x_0 \sim \mathcal{N}(0, I)$ is sampled from a standard normal distribution, and the linear interpolation path is defined as:
\begin{equation}
    x_t = t \cdot x_1 + (1 - t) \cdot x_0, \quad t \in [0, 1],
\end{equation}
with corresponding velocity field $u_t = x_1 - x_0$. The model is trained to predict the velocity field $v_\theta(x_t, t)$ with MSE loss:
\begin{equation}
    \mathcal{L}_\mathrm{flow} = \mathbb{E}_{x_0, x_1, t}\bigl[\| v_\theta(x_t, t) - u_t \|^2\bigr],
\end{equation}
where $t$ is sampled from a Logit-Normal distribution ($\mu{=}0, \sigma{=}1$) to increase sampling density at intermediate timesteps where the velocity field is most informative.

The full generation loss is:
\begin{equation}
    \mathcal{L}_\mathrm{gen} = \lambda_\mathrm{ntp} \cdot \mathcal{L}_\mathrm{NTP} + \lambda_\mathrm{flow} \cdot \mathcal{L}_\mathrm{flow},
\end{equation}
where $\lambda_\mathrm{ntp}{=}1.0$ and $\lambda_\mathrm{flow}{=}0.8$ across all stages; $\mathcal{L}_\mathrm{NTP}$ is next-token prediction loss for text tokens. For pure understanding tasks only $\mathcal{L}_\mathrm{NTP}$ is computed.

At inference, we start from $x_0 \sim \mathcal{N}(0, I)$ and integrate along the learned velocity field with the Euler ODE solver in $N_\mathrm{step}{=}50$ steps, with Classifier-Free Guidance (CFG):
\begin{equation}
    \hat{v} = v_\mathrm{uncond} + s \cdot (v_\mathrm{cond} - v_\mathrm{uncond}),
\end{equation}
where $s{=}3.0$ is the guidance scale. During training, condition dropout (probability 10\%) replaces the conditioning tokens with null embeddings to enable CFG at inference. The generated latent is decoded by the corresponding VAE decoder. Time shifting~\cite{showo2} with factor 3.0 is applied for improved sample quality.

\noindent\textbf{Motion Flow Head.}
The motion flow head consists of $N_d{=}6$ Modulated Attention Blocks with hidden dimension $d_h{=}1536$ and 16 attention heads, followed by a MotionFinalLayer. Timestep conditioning is injected via a sinusoidal timestep embedding $\mathbf{c}_t \in \mathbb{R}^{d_h}$ through Adaptive Layer Normalization (AdaLN): each block modulates the hidden states using shift and scale parameters predicted from $\mathbf{c}_t$. The output layer's weights and biases are \textbf{zero-initialized} to ensure the flow head produces near-zero velocity predictions at the start of training, stabilizing the early training dynamics.

\subsection{Joint-Level Auxiliary Supervision}

For the Vision-to-Motion task, we additionally apply a fine-grained joint reconstruction loss on the decoded motion features. Given the CMA-VAE decoded output $\hat{m} \in \mathbb{R}^{T \times 269}$, a SmoothL1 loss is computed on the first 67 dimensions (encoding root position, root height, and relative joint positions) against the ground truth:
\begin{equation}
    \mathcal{L}_\mathrm{joint} = \mathrm{SmoothL1}(\hat{m}_{1:67},\, m_{1:67}).
\end{equation}
This provides dense per-joint spatial supervision that complements the flow matching objective $\mathcal{L}_\mathrm{flow}$, encouraging geometrically precise joint localization. The frozen pose-aware vision backbone extracts features; only the motion branch and LoRA parameters receive gradients from $\mathcal{L}_\mathrm{joint}$. This auxiliary loss is applied in all training stages involving the Vision-to-Motion task (Stages~2--3), with loss weight $\lambda_\mathrm{joint}{=}0.1$ and a linear warm-up over the first 5000 steps.

\section{Training Pipeline, Data Construction, and Evaluation}
\label{sec:suppl_train}

\subsection{Multi-Stage Training Configuration}

\begin{table}[htbp]
\centering
\renewcommand\arraystretch{1.1}
\caption{Multi-stage training configuration. Each stage builds on parameters from the previous stage.}
\vspace{-1mm}
\label{tab:train_stages}
\resizebox{1.0\linewidth}{!}{%
\begin{tabular}{@{}lllccl@{}}
\toprule
\textbf{Stage} & \textbf{Name} & \textbf{Tasks} & \textbf{Steps} & \textbf{LR} & \textbf{Trainable Params} \\
\midrule
CMA-VAE & DPA Pre-training & Recon.\ + DPA & 210k & 1e-4 & CMA-VAE full ($\sim$45M) \\
Stage 0 & LRA Pre-training & M2M & 80k & 6e-5 & Embedder, FlowHead, LoRA-B \\
Stage 1a & Motion-Text Warmup & T2M & 40k & 6e-5 & All new motion components \\
Stage 1b & Motion-Text Align. & T2M+M2T+Pred+Edit & 130k & 3e-5 & All new motion components \\
Stage 2 & Cross-modal Ext. & +V2M+V2T+MGIE & 130k & 4e-5 & Motion + image components \\
Stage 3 & Full Multi-task FT & All tasks & 200k & 3e-5 & All (LLM partial unfreeze) \\
\bottomrule
\end{tabular}%
}
\vspace{-2mm}
\end{table}

\begin{figure*}[htbp]
    \centering
    \includegraphics[width=\linewidth]{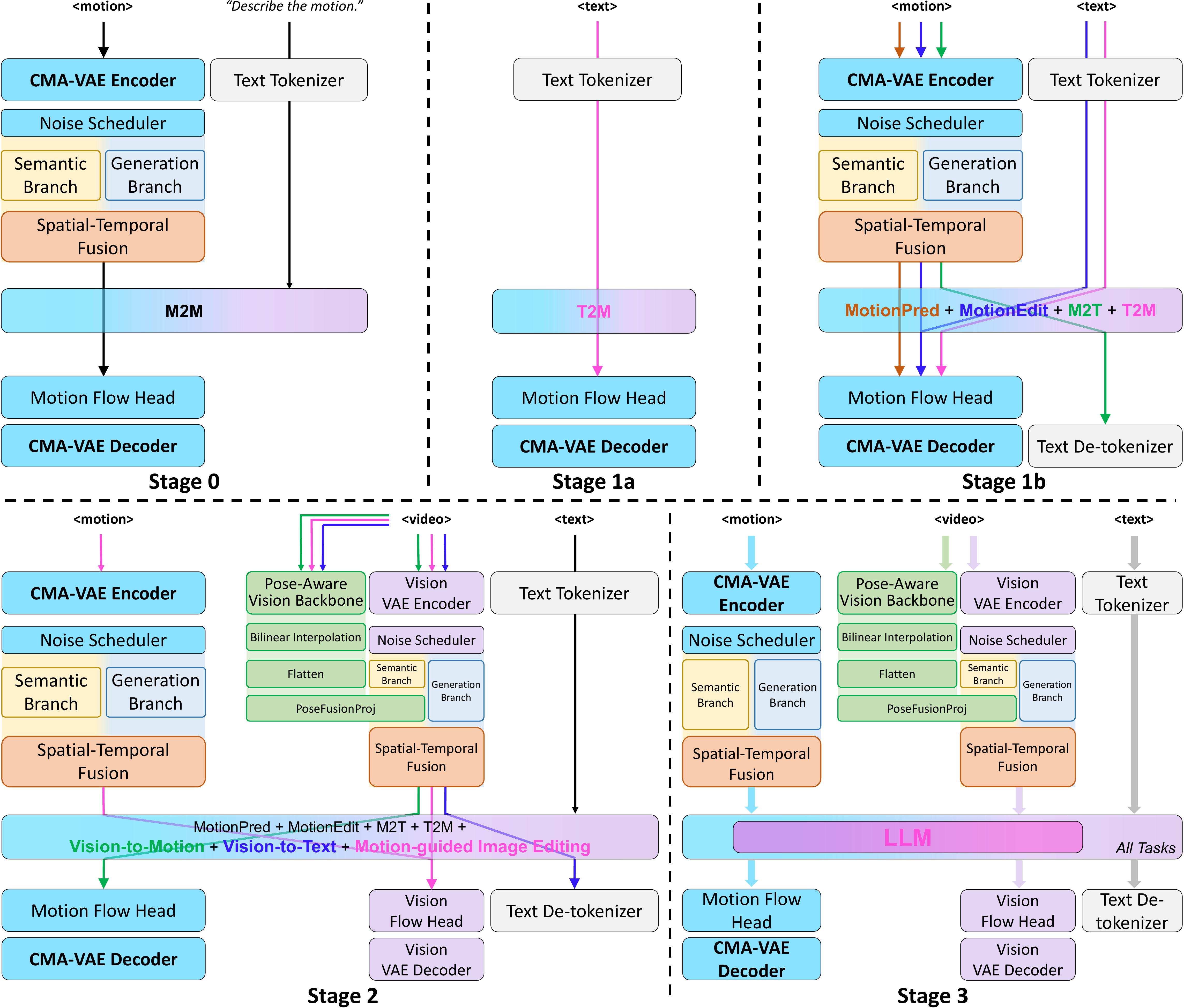}
    \caption{Multi-stage training pipeline of UniMotion. \textbf{(a) Stage~0 -- LRA Pre-training}: the LLM is frozen; motion components are trained via M2M self-reconstruction with information bottlenecks to calibrate the motion pathway. \textbf{(b) Stage~1a -- Motion-Text Warmup}: basic T2M generation capability is established using concise fixed prompt templates. \textbf{(c) Stage~1b -- Motion-Text Alignment}: T2M, M2T, prediction, and editing are jointly trained with diverse instruction templates to achieve bidirectional motion-language alignment. \textbf{(d) Stage~2 -- Cross-modal Extension}: vision tasks (V2M, V2T, MGIE) are added, activating the image pathway and pose-aware backbone to achieve tri-modal coverage. \textbf{(e) Stage~3 -- Full Multi-task Fine-tuning}: all parameters, including the partially unfrozen LLM backbone, are jointly optimized across all seven tasks for global cross-modal integration. CMA-VAE pre-training precedes Stage~0 as an independent phase.}
    \label{fig:suppl_training}
    \vspace{-1mm}
\end{figure*}

Table~\ref{tab:train_stages} summarizes the complete training pipeline. All stages use AdamW optimizer ($\beta_1{=}0.9$, $\beta_2{=}0.999$, weight decay$=$0, $\epsilon{=}$1e-8) with bf16 mixed precision and constant-with-warmup LR scheduling. Training is conducted on 4$\times$A6000 GPUs.
Figure~\ref{fig:suppl_training} provides a visual overview of this progressive pipeline, illustrating which modules are trainable or frozen at each stage and how supervision signals evolve across stages.

\noindent\textbf{CMA-VAE Pre-training.}
Full CMA-VAE training with Dual-Posterior KL Alignment on HumanML3D, MotionFix, and H36M. The DPA alignment loss $\mathcal{L}_\mathrm{align}$ is applied only to samples with paired images (H36M), using a linear warm-up schedule. For HumanML3D and MotionFix samples without paired images, only $\mathcal{L}_\mathrm{recon}$ and $\mathcal{L}_\mathrm{KL}$ are applied.

\noindent\textbf{Stage 0 -- LRA Pre-training.}
M2M self-reconstruction only. To prevent trivial shortcut learning and improve generalization, we apply a lightweight information bottleneck to the conditioning input: temporal subsampling (randomly retaining 20--50\% of frames), feature dropout ($p{=}0.15$), and low-level Gaussian perturbation ($\sigma{=}0.02$). The dual-path embedder, motion flow head, and modality-routed LoRA-B (Motion branch) participate in training. Training data includes HumanML3D, H36M, and MotionFix motion sequences. Warmup steps: 2000.

\noindent\textbf{Stage 1a -- Motion-Text Warmup.}
T2M using concise fixed prompt templates, establishing basic text-to-motion generation capability. Warmup steps: 2000.

\noindent\textbf{Stage 1b -- Motion-Text Alignment.}
Joint training of T2M + M2T + Motion Prediction + Motion Editing with rich and diverse instruction templates. Warmup steps: 4000.

\noindent\textbf{Stage 2 -- Cross-modal Extension.}
Three vision-modality tasks are added: Vision-to-Motion, Vision-to-Text, and Motion-guided Image Editing (MGIE). Visual inputs are processed via WAN2.1 VAE~\cite{wan2025wan} and Show-o2's native image pipeline, with the pose-aware vision backbone additionally activated. Stage 2 proceeds in two phases: first img2pose + img2text, then MGIE + img2pose + img2text. Warmup steps: 2000.

\noindent\textbf{Stage 3 -- Full Multi-task Fine-tuning.}
All tasks are jointly trained with partial LLM backbone unfreezing in the final full multi-task stage. Warmup steps: 4000.

\subsection{Cross-Modal Task Data Construction}
\label{sec:suppl_data}

This section details the data construction for cross-modal tasks in the later training stages (Stages 2--3).

\noindent\textbf{Vision-to-Motion (i2m) and Vision-to-Text (i2t).}
Both tasks are built on Human3.6M visual inputs and corresponding 269-dim motion representations, using online sample construction. In the current H3.6M setup, each sample uses one reference frame, with unified resolution (432$\times$432) and normalization applied before training. Task-specific quality filtering ensures i2m covers the broadest possible visual-motion samples, while i2t additionally requires available text supervision.

\noindent\textbf{Motion-guided Image Editing (MGIE).}
MGIE targets generation from ``source image + reference motion + text instruction'' to produce a target image. We perform temporal pairing within the same video sequence:
\begin{itemize}
    \item Candidate target frames are sampled using a discrete time interval set ($\{5, 8, 10\}$ frames), covering short-to-medium motion changes.
    \item A pose difference threshold (based on $L_2$ distance in 269-dim motion space) filters out pairs with minimal changes.
    \item At most one target frame is retained per source frame to control sample correlation.
\end{itemize}
Under this configuration, we obtain approximately 134k training pairs and 40.6k test pairs.

\noindent\textbf{Instruction templates.}
Diverse instruction template sets are constructed for each task, with random sampling during training. MGIE instructions explicitly constrain ``executing motion editing while preserving the scene/subject,'' reinforcing decoupled modeling of motion change and appearance preservation.

\subsection{Evaluation Protocols}
\label{sec:suppl_eval}

We detail the evaluation protocols for all tasks to ensure reproducibility.

\noindent\textbf{Text-to-Motion (T2M).}
Evaluated on the HumanML3D test split (4,646 samples). We compute FID, R-Precision (R@1/2/3), MMDist, and Diversity following the standard protocol~\cite{humanml3d}. Motion features are extracted using the text-motion feature extractor from Guo~\emph{et al.}~\cite{guo2022generating}. Each metric is averaged over 20 runs with different random seeds.

\noindent\textbf{Motion-to-Text (M2T).}
Evaluated on HumanML3D test split following~\cite{tm2t}. We report retrieval metrics (R@1/3, MMDist) and caption quality metrics (Bleu@1/4, Rouge-L, CIDEr, BertScore).

\noindent\textbf{Motion Prediction.}
Evaluated on the AMASS subset following MotionGPT~\cite{motiongpt}. Given the first half of a motion, the model predicts the remaining half. We report FID, ADE (Average Displacement Error), and FDE (Final Displacement Error).

\noindent\textbf{Motion Editing.}
Evaluated on the MotionFix test split (2,847 samples). Given a source motion and a text editing instruction, the model generates the edited motion. We report FID and generated-to-target retrieval precision (R@1/3) following~\cite{motionfix}. Motion Editing FID is generally lower than T2M FID because the editing task has dual constraints (source motion + text instruction), producing distributions closer to the ground truth.

\noindent\textbf{Vision-to-Motion.}
Evaluated on Human3.6M test split following~\cite{unipose}. We report MPJPE and PA-MPJPE. The same data splits and evaluation code as UniPose are used for fair comparison.

\noindent\textbf{Vision-to-Text.}
Evaluated on Human3.6M test split. We report BLEU-4, ROUGE-L, and METEOR for pose description quality. All baselines (UniPose, Show-o2, Qwen-2.5-VL) are evaluated on the same split with the same tokenizer settings.

\noindent\textbf{Motion-guided Image Editing (MGIE).}
Evaluated on 40.6k test pairs constructed from H3.6M and 3DPW~\cite{3dpw}. We report FID (Inception-v3 features), CLIP Score (text-image alignment), and Motion Accuracy (\textbf{Mot.Acc}). Specifically, Mot.Acc measures the pose condition execution success rate rather than a classification accuracy. For each generated image, we use HMR2.0~\cite{hmr2} to extract its human pose and calculate the Procrustes-Aligned Mean Per Joint Position Error (PA-MPJPE) against the target reference motion. A generation is considered a successful ``hit'' if the PA-MPJPE is less than or equal to a strict threshold ($100.0$\,mm). Mot.Acc is defined as the percentage of hits across all samples, directly reflecting how reliably the generated image conforms to the given motion condition. Training and evaluation sets have no overlap.

\section{Limitations and Broader Impact}
\label{sec:suppl_limit}

\noindent\textbf{Limitations.}
UniMotion inherits the computational overhead of a 1.5B-parameter backbone, which may limit deployment in resource-constrained settings. The pose-aware vision backbone relies on a frozen pretrained human body encoder~\cite{tokenhmr}; robustness to severe occlusion, camera motion, and diverse in-the-wild scenarios remains to be thoroughly validated, as the visual-motion alignment is primarily established on indoor datasets (Human3.6M). The current Vision-to-Motion evaluation on Human3.6M uses frame-based visual inputs; extending the same interface to richer video-level temporal reasoning is a promising direction. We hope UniMotion opens a practical direction toward motion-aware multimodal intelligence.

\noindent\textbf{Broader impact.}
UniMotion's unified motion-language-vision interface can benefit animation, AR/VR, robotics, and medical rehabilitation. Potential risks include synthetic media misuse and privacy concerns if trained on identifiable human videos. We commit to clear dataset licensing, anonymization, and responsible use.